%% file: aaai_2016_sgnht.tex
\def\year{2016}
\documentclass[letterpaper]{article}
\include{subtex/mlVecMat}

\usepackage{aaai16}
\usepackage{times}
\usepackage{helvet}
\usepackage{courier}

\usepackage{marvosym}
\newcommand{\envelope}{\raisebox{-.5pt}{\scalebox{1.05}{\Letter}}\kern1.0pt}

\title{High-Order Stochastic Gradient Thermostats for \\Bayesian Learning of Deep Models
	}
\author{Chunyuan Li$^{1}$, Changyou Chen$^{1}$, Kai Fan$^{2}$ and Lawrence Carin$^{1}$\\
	$^{1}$Department of Electrical and Computer Engineering,  Duke University\\
	$^{2}$Computational Biology and Bioinformatics, Duke University\\
	{\footnotesize
		\href{mailto:chunyuan.li@duke.edu}{  \;\;\;\;\nolinkurl{chunyuan.li@duke.edu,}  } 
		\href{mailto:cchangyou@gmail.com}{\nolinkurl{cchangyou@gmail.com,}  } 
		\href{mailto:kai.fan@stat.duke.edu}{\nolinkurl{kai.fan@stat.duke.edu,}  } 
		\href{mailto:lcarin@duke.edu}{\nolinkurl{lcarin@duke.edu} }
	}
}
\begin{document}
% \title{High-Order Stochastic Gradient Thermostats for \\Bayesian Learning of Deep Models}
%\author{AAAI Press\\
%Association for the Advancement of Artificial Intelligence\\
%2275 East Bayshore Road, Suite 160\\
%Palo Alto, California 94303\\
%}
\maketitle
\begin{abstract}
\begin{quote}	
Learning in deep models using Bayesian methods has generated significant attention recently. This is largely because of the feasibility of modern Bayesian methods to yield scalable learning and inference, while maintaining a measure of uncertainty in the model parameters. Stochastic gradient MCMC algorithms (SG-MCMC) are a family of diffusion-based sampling methods for large-scale Bayesian learning. In SG-MCMC, multivariate stochastic gradient thermostats (mSGNHT) augment each parameter of interest, with a momentum and a thermostat variable to maintain stationary distributions as target posterior distributions. As the number of variables in a continuous-time diffusion increases, its numerical approximation error becomes a practical bottleneck, so better use of a numerical integrator is desirable. To this end, we propose use of an efficient symmetric splitting integrator in mSGNHT, instead of the traditional Euler integrator. We demonstrate that the proposed scheme is more accurate, robust, and converges faster. These properties are demonstrated to be desirable in Bayesian deep learning. Extensive experiments on two canonical models and their deep extensions demonstrate that the proposed scheme improves general Bayesian posterior sampling, particularly for deep models.
\end{quote}
\end{abstract}

% Bayesian deep models have recently regained popularity, largely because of the feasibility of scalable learning and inference. Stochastic gradient MCMC algorithms (SG-MCMC) are a family of diffusion-based sampling methods for large-scale Bayesian learning. In SG-MCMC, multivariate stochastic gradient thermostats (mSGNHT) augments each parameter of interest, with a momentum and a thermostat variable to maintain stationary distributions as target posterior distributions. As the number of variables in a continuous-time diffusion increases, its numerical approximation error becomes a practical bottleneck, so better use of a numerical integrator is favorable. To this end, we propose to use an efficient symmetric splitting integrator in mSGNHT, instead of the traditional Euler integrator. We justify that the proposed scheme is more accurate, robust, and converges faster. These properties are further articulated to be desirable in Bayesian deep learning. Extensive experiments on two canonical Bayesian models and their deep extensions demonstrate that the proposed scheme improves general Bayesian posterior sampling, particularly for Bayesian deep models.

\section{Introduction}

The ability to learn abstract representations that support generalization to novel instances lies at the core of many problems in machine learning and computer vision. Human learners often can grasp concepts at multiple levels of abstraction from training examples,  and make meaningful generalizations~\cite{kemp2007learning}. Intuitively, appropriately employed prior knowledge and hierarchical reasoning are necessary in this task. Bayesian learning and inference applied to deep models may naturally possess such characterization, and potentially could take a step towards this ability \cite{salakhutdinov2013learning}.

% In typical applications of topic modeling, sophisticated algorithms needs to be carefully designed to discover desirable patterns in a corpus.
% BDM are derived by composing deep networks in a Bayesian flavor. There are two main categories. 

Deep models come in two broad categories. The first uses stochastic hidden layers, typically deep latent variable models. This includes the deep sigmoid belief network~\cite{mnih2014neural,gan2015learning}, the variational auto-encoder~\cite{kingma2013auto}, and many others~\cite{ranganath2015deep,gan2015deep,pu2015generative}. The second category of deep models uses deterministic  hidden layers.  While the stochastic hidden units of the first category make this class of models naturally amenable to Bayesian learning (see the above references), for the second category appropriate priors on the weights of networks may be employed to consider weight uncertainty. Previous work has applied Bayesian methods to neural networks~\cite{mackay1992practical,neal1995bayesian}, including feedforward neural networks~\cite{blundell2015weight,hernandez2015probabilistic,korattikara2015bayesian} and convolutional neural networks~\cite{Gal2015Bayesian,li2016psgld}. Deep learning may often be interpreted as a stacking of such neural networks.

Bayesian learning and inference methods have generated significant recent research activity.
Stochastic gradient Markov Chain Monte Carlo (SG-MCMC) methods~\cite{welling2011bayesian,ChenFG:ICML14,DingFBCSN:NIPS14} are a family of It\^{o} diffusion based algorithms that can efficiently sample target distributions, and can be applied to large datasets. In these algorithms, two approximations are made~\cite{chen2015integrator}. $(\RN{1})$ For \textit{practical scalability}, stochastic gradients from minibatches of data are used to estimate the true gradient; $(\RN{2})$ For \textit{numerical feasibility}, a numerical integration with small step is used to solve the corresponding It\^{o} diffusion (a continuous-time Markovian process).

The first attempt at SG-MCMC was the Stochastic Gradient Langevin Dynamics (SGLD)~\cite{welling2011bayesian} . It is based on 1st-order Langevin dynamics. The Stochastic Gradient {H}amiltonian {M}onte {C}arlo (SGHMC)~\cite{ChenFG:ICML14} extends SGLD with 2nd-order Langevin dynamics, where momentum variables are introduced into the system. In an attempt to address the problem of estimating stochastic gradient noise, the Stochastic Gradient No\'{s}e-Hoover Thermostat (SGNHT)~\cite{DingFBCSN:NIPS14} was proposed, with one additional global thermostat variable. To further improve the efficiency of the SGNHT, a multivariate version of SGNHT (mSGNHT) was proposed by introducing multiple thermostat variables instead of a single one~\cite{DingFBCSN:NIPS14}. It was shown that mSGNHT provides more adaptivity than SGNHT \cite{gan2015scalable}.

By examining training with SG-MCMC algorithms, we note two issues.
$(\RN{1})$ as more variables are introduced, an accurate numerical method becomes more critical; and
$(\RN{2})$ gradients in deep models often suffer from the vanishing/exploding problem~\cite{bengio1994learning}, which makes choosing a proper stepszie difficult in SG-MCMC. In this paper, we mitigate these concerns by utilizing a more accurate numerical integrator, the symmetric splitting integrator (SSI), to reduce discretization errors in mSGNHT. Furthermore, since SSI is more robust with respect to stepsizes than the default Euler integrator, it allows one to choose an appropriate stepsize much more easily. We justify that the Euler integrator used in mSGNHT is 1st-order, while the SSI is 2nd-order. Borrowing tools from \cite{chen2015integrator}, we show that mSGNHT with SSI (mSGNHT-S) converges faster and more accurately than mSGNHT with a Euler integrator (mSGNHT-E). Experiments across a wide range of model types demonstrate the utility of this method. Specifically, we consider latent Dirichlet allocation, logistic regression, deep neural networks and deep Poisson factor analysis. 
%These experiments demonstrate the superiority of mSGNHT-S to mSGNHT-E.

% By borrowing attributes of both categories, a novel Temporal Poission Dynamic Model is designed and tested.

\section{Background}

% It\^{o} diffusion is the core of general SG-MCMC. Taking mSGNHT as an example,  we review this concept, as well as the popular Euler integrator for numerically sampling from an It\^{o} diffusion in SG-MCMC. 

\subsection{It\^{o} Diffusion}

It\'{o} diffusion is a stochastic differential equation (SDE) defined as: 
\begin{align}
\mathrm{d}\Xmat_t &= F(\Xmat_t)\mathrm{d}t + \sigma(\Xmat_t)\mathrm{d}W_t~, \label{eq:itodiffusion}
\end{align}
where $\Xmat_t \in \R^n$, $W_t$ is Brownian motion, and $t$ is the time index.
Functions $F: \R^n \to \R^n$ and $\sigma: \R^n \rightarrow \R^{n\times m}$ are assumed 
to satisfy the usual Lipschitz continuity condition~\cite{knapp2005basic}. It has been shown that by designing appropriate functions $F$ and $\sigma$, the stationary distribution, $\rho(\Xmat)$, of the It\^{o} diffusion \eqref{eq:itodiffusion} has a 
marginal distribution that is equal to the posterior distribution of interest \cite{chen2015integrator,ma2015complete}.

To formulate mSGNHT~\cite{gan2015scalable,DingFBCSN:NIPS14} into the It\'{o} diffusion \eqref{eq:itodiffusion}, 
let $\Xmat = (\thetav, \pv, \xiv)$, where $\thetav \in \R^n$ are the model parameters, $\pv \in \R^n$ are momentums, 
and $\xiv \in \R^{n}$ represent the thermostats~\cite{gan2015scalable}\footnote{$\Xmat$ now is in $\R^{3n}$; for conciseness, we do not re-define the dimension for $\Xmat$ in \eqref{eq:itodiffusion}. }. 
%In addition, we restrict $\xiv$ to be a diagonal matrix for computationally feasibility.
Define $U \triangleq -\left(\sum_{i=1}^N\log p(d_i | \thetav) + \log p(\thetav)\right)$ as the unnormalized
negative log-posterior, where $\{d_i\}$ represents the $i$th data sample, $p(d_i | \thetav)$ the corresponding likelihood, and $p(\thetav)$ the prior. 
For some constant $D > 0$, the mSGNHT in \cite{gan2015scalable} is shown to be in a form of It\^{o} diffusion, with
\begin{align}\label{eq:msgnht_decom}
F=  \begin{bmatrix}
\pv  \\
- \xiv \odot \pv -\nabla_{\thetav} U  \\
  \pv \odot \pv - 1
\end{bmatrix},~~
\sigma = \sqrt{2D}\begin{bmatrix}
{\bf 0} & {\bf 0}  & {\bf 0} \\
{\bf 0} & \Imat_n  & {\bf 0} \\
{\bf 0} & {\bf 0} & {\bf 0}
\end{bmatrix}~,
\end{align}
where $\odot$ represents element-wise product, and $\Imat_n$ is the $n\times n$ identity matrix.
Based on the Fokker-Planck equation~\cite{Risken:FPE89}, the marginal stationary distribution over $\thetav$
can be shown to be $\rho(\thetav) \propto \exp(-U(\thetav))$, the posterior distribution we are interested in.
%\begin{align}
%& P(\thetav, \pv, \xiv) \propto \\ & \exp { \Big ( -U(\thetav) - \frac{1}{2}\pv^T\pv + \frac{1}{2} \text{tr} \big \{ (\xiv - D)^\top  (\xiv - D) \big \}  \Big ) }~. \nonumber
%\end{align}
%%
%By marginalizing over $\thetav$, we obtain $\rho(\thetav) \propto \exp(-U(\thetav))$.
%
%The SGNHT \cite{DingFBCSN:NIPS14} is based on the No\'{s}e-Hoover thermostat defined as:
%\begin{align}\label{eq:NHT}
%\left\{\begin{array}{ll}
%\mathrm{d}\thetav &= \pv \mathrm{d}t \\
%\mathrm{d}\pv &= -\nabla_{\thetav} U(\thetav) \mathrm{d}t -  \text{diag} (\xiv) \pv \mathrm{d}t + \sqrt{2D} \mathrm{d}W \\
%\mathrm{d}\xiv &=  \left( \text{diag} ( \pv \odot \pv ) - \Imat \right)~\mathrm{d}t,
%\end{array} \right.
%\end{align}
%If $D$ is independent of $\thetav$ and $\pv$, it can also be shown that the equilibrium distribution
%of \eqref{eq:NHT} is \cite{DingFBCSN:NIPS14}:

\vspace{0mm}
\subsection{Euler Integrator}

The continuous-time diffusion in \eqref{eq:itodiffusion} cannot be solved explicitly in general. As a result,
numerical methods are required in SG-MCMCs to generate approximate samples. The standard numerical method 
used in SG-MCMC is the Euler integrator, which generates samples sequentially from a discrete-time 
approximation of \eqref{eq:itodiffusion}. Specifically, conditioned on the current sample $\Xmat_t$
and step size $h$, the next sample at time $t+1$ is generated via the rule:
\begin{align*}
	\Xmat_{t+1} = \Xmat_{t} + F(\Xmat_t) h + \sigma(\Xmat_t) \zetav_{t+1}, \hspace{0.2cm}\zeta_{t+1} \sim \mathcal{N}(\mathbf{0}, h\mathbf{I}_n)~.
\end{align*}

In the case of mSGNHT, in each step, a stochastic gradient from a minibatch is used instead of the full gradient.
We thus approximate $U$ with 
$\tilde{U}_t \triangleq -\left(\frac{N}{|\mathcal{S}_t|}\sum_{i\in \mathcal{S}_t}\log p(d_i | \thetav) + \log p(\thetav)\right)$
for the $t$-th iteration, where $\mathcal{S}_t \subset \{1, 2, \cdots, N\}$, and $|\cdot|$ is the cardinality of a set\footnote{We
write $F(\Xmat_t)$ from \eqref{eq:msgnht_decom} as $\tilde{F}(\Xmat_t)$ if a stochastic gradient is used in the rest of the paper.}.
This results in the following sampling rules:
\begin{align*}
	&\left\{\begin{array}{ll}
	\thetav_{t+1} &= \thetav_t + \pv_t h \\
	\pv_{t+1} &= \pv_t - \nabla_{\thetav}\tilde{U}_t(\thetav_{t+1}) h - \text{diag} ( \xiv_t ) \pv_t h + \sqrt{2D} \zetav_{t+1} \\
	\xiv_{t+1} &= \xiv_t + \left(  \pv_{t+1} \odot \pv_{t+1} - 1\right) h
	\end{array}\right.
\end{align*}

%\begin{algorithm}
%	\SetKwInOut{Input}{Input}
%	\caption{Stochastic Gradient Nos\'{e}-Hoover Thermostats}
%	\Input{Parameters $h, D$.}
%	Initialize $\thetav_{0} \in \Re^n$, $\pv_{0} \sim \Ncal(0,\Imat)$, and $\xi_{0}=D$ \;
%	\For {$l = 1, 2, \ldots $} {
%		Evaluate $\nabla \tilde{U}_l(\thetav_{(l-1)h})$ from the $l$-th minibatch \;
%		$\pv_{lh} = \pv_{(l-1)h} - \xi_{(l-1)h} \pv_{(l-1)h} h - \nabla \tilde{U}_l(\thetav_{(l-1)h}) h + \sqrt{2Dh}\Ncal(0, I)$\;
%		$\thetav_{lh} = \thetav_{(l-1)h} + \pv_{lh} h$\;
%		$\xi_{lh} = \xi_{(l-1)h} + (\frac{1}{n} \pv_{lh}^{\top} \pv_{lh} - 1) h $\;
%	}
%	\label{alg:sgnht}
%\end{algorithm}

\section{Symmetric Splitting Integrator for mSGNHT}

The SSI has been studied in statistical physics \cite{LeimkuhlerM:AMRE13,leimkuhler2015adaptive}.
It generalizes the idea of the {\em leap-frog} integrator used in the Hamiltonian Monte Carlo \cite{neal2011mcmc} from 
the partial differential equation setting to the SDE setting.
It was not until recently that SSI was introduced into machine learning to obtain a more accurate SGHMC algorithm \cite{chen2015integrator}.
We adopt the idea and generalize it in this paper for mSGNHT.

The idea of SSI is to split the intractable SDE, {\it i.e.}, \eqref{eq:itodiffusion}, into several sub-SDEs such that each sub-SDE
can be solved analytically. For the mSGNHT represented in \eqref{eq:msgnht_decom}, it is readily split into the following sub-SDEs:
\begin{align}\label{eq:split_msgnht}
	&A: \left\{\begin{array}{ll}
	\mathrm{d}\thetav &= \pv \mathrm{d}t \\
	\mathrm{d}\pv &= 0 \\
	\mathrm{d}\xiv &= \left( \pv \odot \pv  - \Imat \right) \mathrm{d}t
	\end{array}\hspace{-1mm},\right. \,
	B: \left\{\begin{array}{ll}
	\mathrm{d}\thetav &= 0 \\
	\mathrm{d}\pv &= - \xiv \odot \pv \mathrm{d}t, \\
	\mathrm{d}\xiv &= 0
	\end{array}\right.\nonumber\\
	&O: \left\{\begin{array}{ll}
	\mathrm{d}\thetav &= 0 \\
	\mathrm{d}\pv &= -\nabla_{\thetav} \tilde{U}_t(\thetav) \mathrm{d}t + \sqrt{2D}\mathrm{d}W. \\
	\mathrm{d}\xiv &= 0
	\end{array}\right.
\end{align}

All the sub-SDEs can be solved analytically, leading to the
following rules to generate samples $\{\thetav_{t+1}, \pv_{t+1}, \xiv_{t+1}\}$ from mSGNHT
for time $(t+1)$:
{\small 
\begin{align*}
	&A: \thetav_{t+1/2} = \pv_t h / 2, \hspace{0.2cm}\xiv_{t+1/2} = \xiv_t + \left( \pv_t \odot \pv_t -  1 \right) h / 2 \rightarrow \\
	&B: \pv_{t+1/3} = \exp( -\xiv_{t+1/2} h / 2) \odot \pv_t \rightarrow \\
	&O:  \pv_{t+2/3} = \pv_{t+1/3} - \nabla_{\thetav} \tilde{U}_t(\thetav_{t+1/2}) h + \sqrt{2D} \zetav_{t+1} \rightarrow \\
	&B: \pv_{t+1} = \exp( -\xiv_{t+1/2} h / 2) \odot \pv_{t+2/3} \rightarrow \\
	&A: \thetav_{t+1} = \pv_{t+1} h / 2, \hspace{0.2cm}\xiv_{t+1} = \xiv_{t+1/2} + \left( \pv_{t+1} \odot \pv_{t+1} - 1 \right) h / 2
\end{align*}
}

From the update equations, SSI performs almost as efficiently as the Euler integrator.
Furthermore, the splitting scheme for \eqref{eq:msgnht_decom} is not unique. However, all of the schemes can be shown to have the same order of accuracy. In the following subsection, we show quantitatively that the SSI is more accurate than the Euler integrator in terms of approximation errors. To get an impression of how the SSI works, we illustrate it with a simple synthetic experiment.

\vspace{-0.05cm}\paragraph{Illustrations with a Double-well Potential}
To illustrate the proposed symmetric splitting scheme and its robustness to stepsize, fast convergence, and accurate parameter approximation, we follow~\cite{DingFBCSN:NIPS14}, and consider the  double-well potential with
\begin{align*}
U(\theta) = (\theta + 4)(\theta + 1)(\theta - 1)(\theta - 3)/14 + 0.5,
\end{align*}
%

%\vspace{-3mm}
\noindent and the target distribution $\rho(\theta) \propto \exp(-U(\theta))$. The unknown noise in the stochastic gradient is simulated as $\nabla \tilde{U}(\theta) h =  \nabla U(\theta) h + \Ncal(0, 2Bh)$, where $B = 1$. No injecting noise is added. We examine a large range of stepsize $h$ from $10^{-3}$ to $0.3$. 

%\begin{figure}
\begin{wrapfigure}{R}{3.8cm}%\vspace{-25pt}
	\vspace{-4mm}
	%\hspace{-2mm}
	\centering
	\hspace{-3mm}\includegraphics[width=4.0cm]{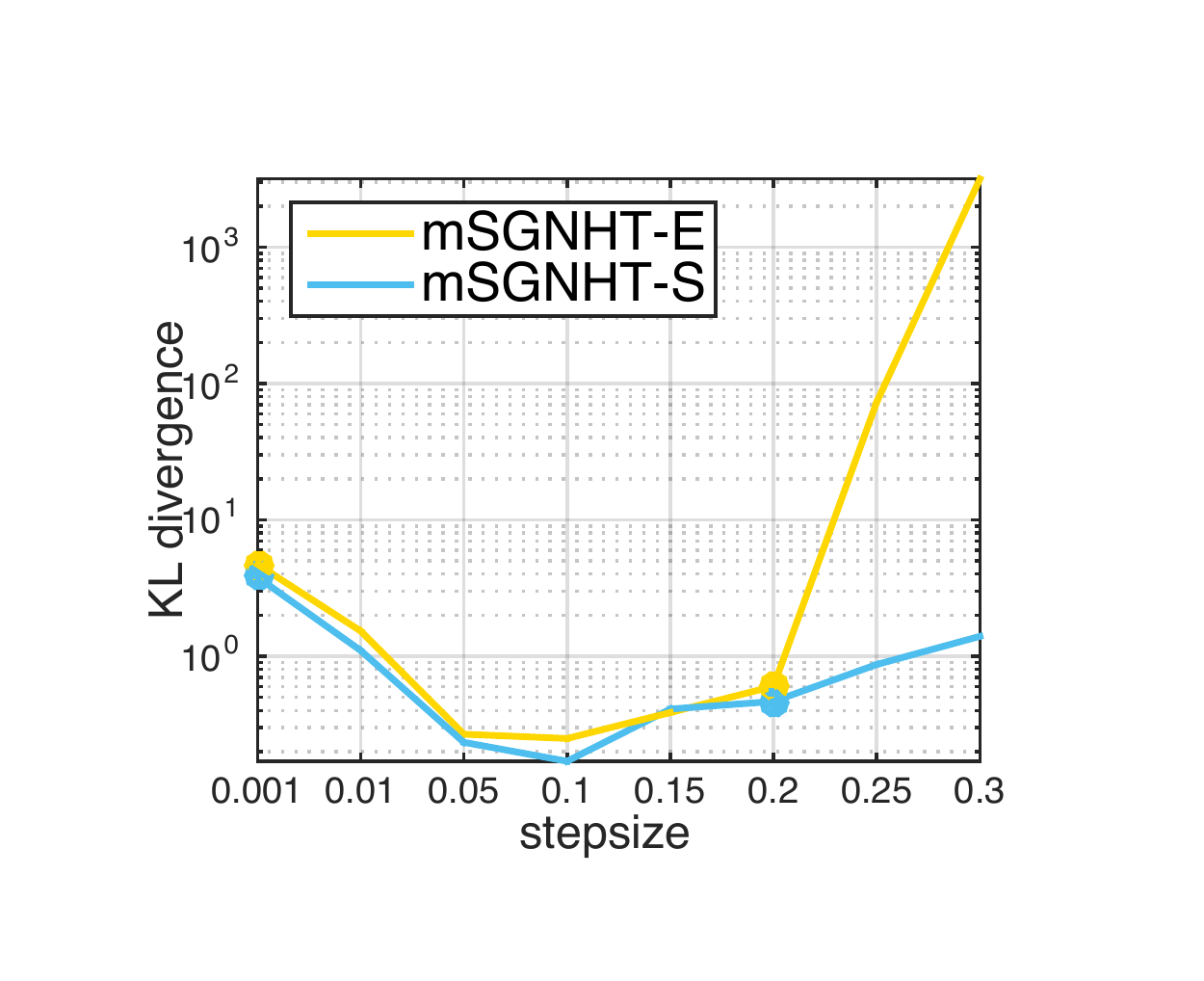}
	\vspace{-2mm}
	\caption{KL divergence for varying stepsize.}
	\label{fig:kl_2wells}
	\vspace{-1em}
\end{wrapfigure}
%\end{figure}

In Fig.~\ref{fig:kl_2wells}, we plot the KL divergences between the true distributions and the estimated density, based on $10^6$ samples, using two types of integrators. mSGNHT-S consistently provides a better approximation. The significant gap at larger stepsize reveals that mSGNHT-S allows large updates.
%\vspace{-0mm}

\begin{figure}[h!]  \tiny \centering
	\vspace{4mm}
	\begin{tabular}{ c c } %\hline
		\begin{minipage}{3.6cm}
			\includegraphics[width=3.6cm]{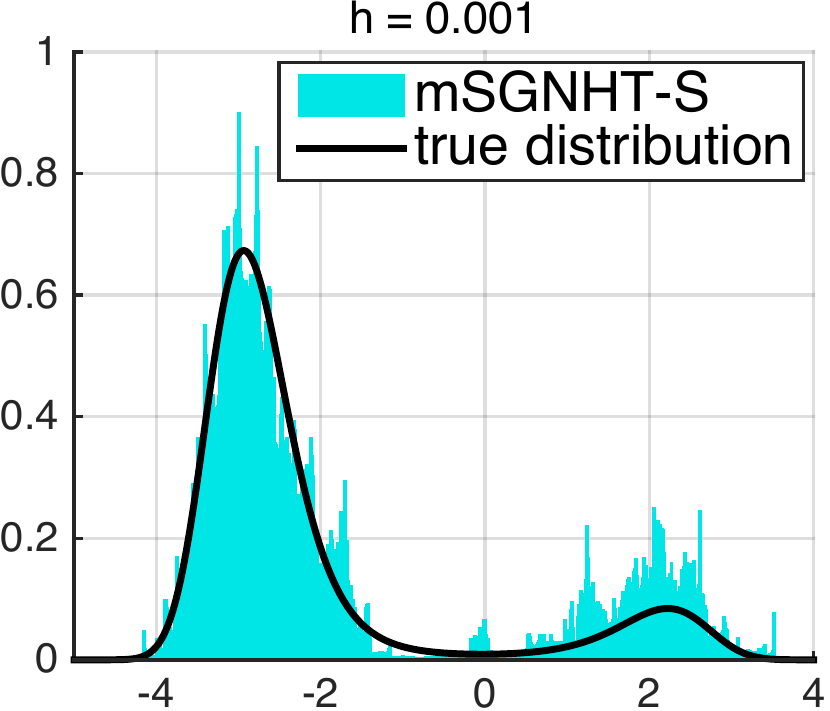} 
		\end{minipage}   &
		\hspace{-4mm}
		\begin{minipage}{3.6cm}
			\includegraphics[width=3.6cm]{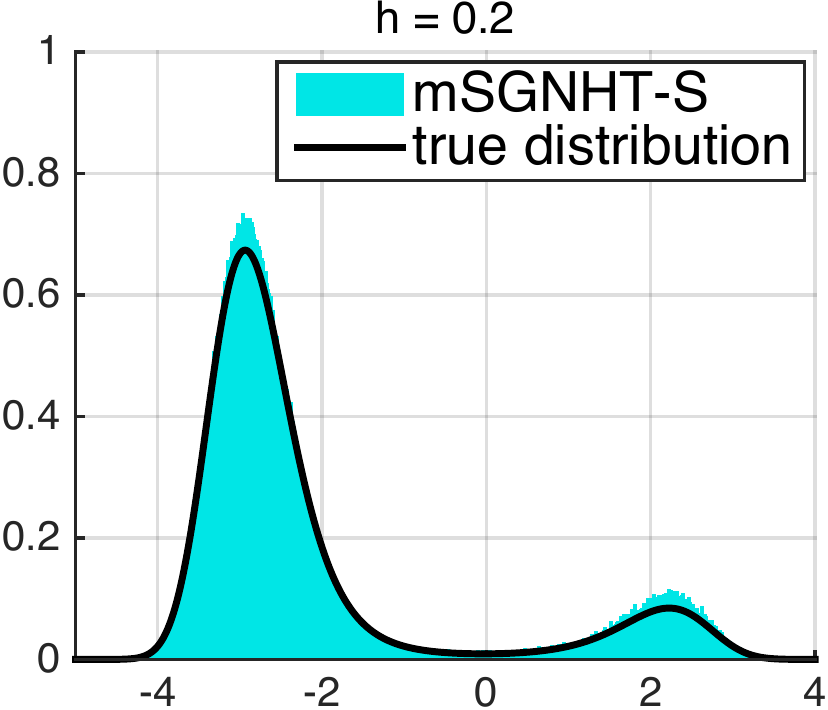} 
		\end{minipage}  \\		
		%\hspace{-4mm}
		\begin{minipage}{3.6cm}
			\includegraphics[width=3.6cm]{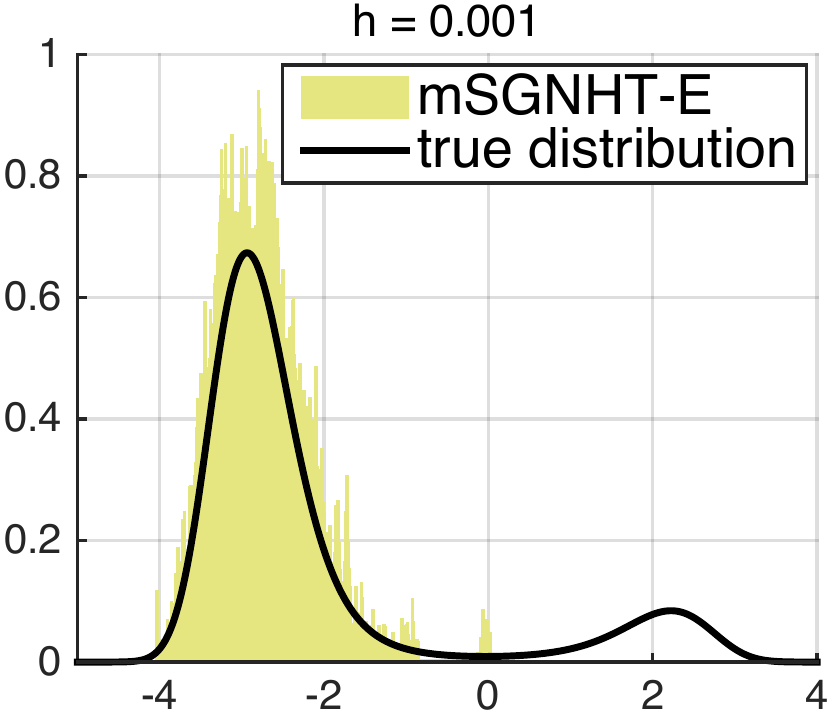} 
		\end{minipage}  &
		\hspace{-3mm}
		\begin{minipage}{3.6cm}
			\includegraphics[width=3.6cm]{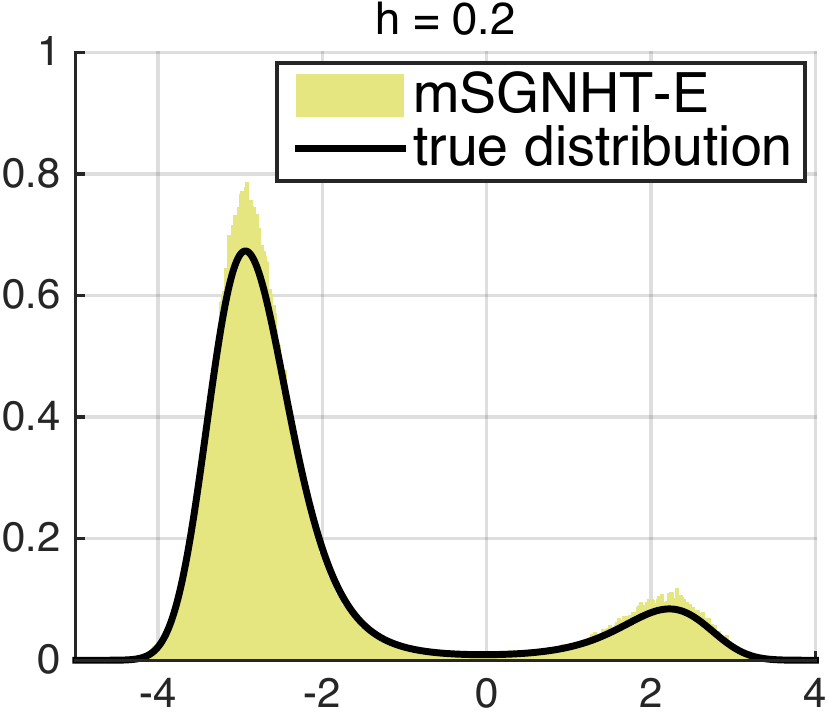} 
		\end{minipage}   
		\vspace{-2mm}\\		
		%\hspace{-4mm}
		
		\begin{minipage}{3.6cm}\vspace{3mm}
			\includegraphics[width=3.6cm]{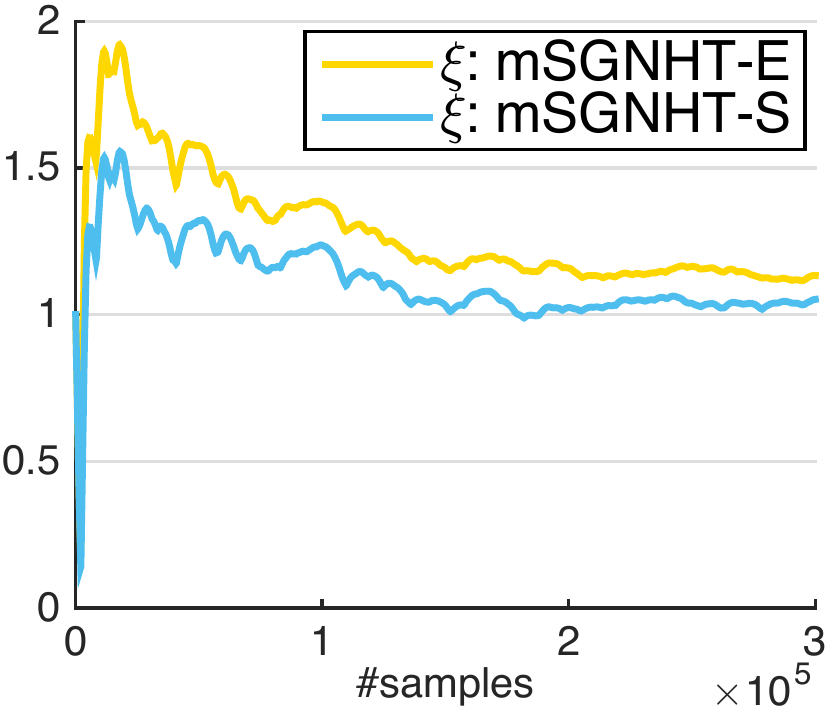} 
		\end{minipage} &
		\hspace{-4mm}
		\begin{minipage}{3.6cm}\vspace{3mm}
			\includegraphics[width=3.6cm]{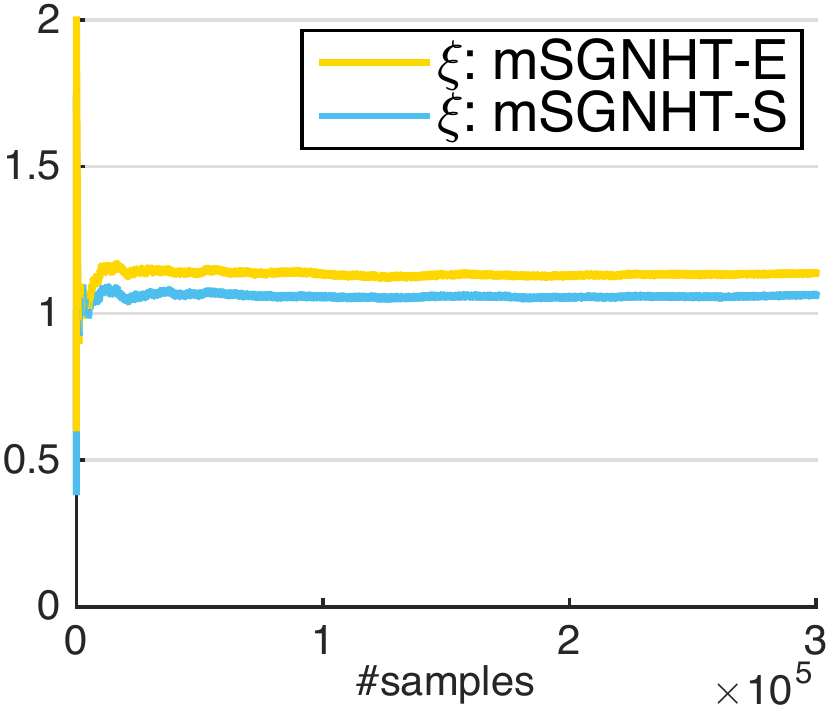} 
		\end{minipage} \\ %\hline
	\end{tabular}

	\caption{Samples of $\rho(\theta)$ with SSI (1st row) and Euler integrator (2nd row),  and the estimated thermostat variable over iterations (3rd row). }	
	\hspace{-4mm}\vspace{-4mm}
	\label{fig:distribution_2wells}	
\end{figure}

Furthermore, we visualize the results of the first $10^5$ samples for $h = 10^{-3}$ and $h = 0.2$ in Fig.~\ref{fig:distribution_2wells}. When the stepsize is too small ($h = 10^{-3}$), conventional mSGNHT-E has not explored the whole parameter space; this is because it converges slower, as shown later.
When the stepsize is large ($h = 0.2$), large numerical error is potentially brought in, and mSGNHT-E over-concentrates on the mode. In both cases, mSGNHT-S approaches the theoretical value of thermostat variable $\xi=1$ more accurately.

\vspace{-0mm}

\subsection{Theoretical Justification}

In~\cite{chen2015integrator}, the authors formally studied the roles of numerical integrators in general SG-MCMCs.
We adopt their framework, and justify the advantage of the proposed scheme for mSGNHT. We first define the
local generator of the SDE \eqref{eq:itodiffusion} at the $t$-th iteration ( \ie replacing the full gradient with the
stochastic gradient from the $t$-th minibatch) as:
{\small\begin{align*}
	\hspace{-0.0cm}\tilde{\mathcal{L}}_tf(\Xmat_t) %&\triangleq \lim_{h \rightarrow 0^{+}} \frac{\mathbb{E}\left[f(\Xmat_{t+h})\right] - f(\Xmat_t)}{h} 
	\!\triangleq\! \left(\tilde{F}_t(\Xmat_t)\! \cdot \!\nabla_{\Xmat} \!+ \!\frac{1}{2}\left(\sigma(\Xmat_t) \sigma(\Xmat_t)^\top\right)\!:\! \nabla_{\Xmat} \nabla_{\Xmat}^\top\right) \!f(\Xmat_t)
\end{align*}}
\noindent 
\!\!where $\av\cdot \bv \triangleq \av^\top\bv$, $\Amat\!:\!\Bmat \triangleq \text{tr}(\Amat^\top \Bmat)$, $f:\mathbb{R}^n \rightarrow \mathbb{R}$
is any twice differentiable function. Based on the definition, according to the Kolmogorov backward equation, 
we have $\mathbb{E}[f(X)] = e^{h \tilde{\mathcal{L}}_t} f(\Xmat)$ where the expectation is taken over the randomness in the diffusion. The operator $e^{h \tilde{\mathcal{L}}_l}$ 
is called the Kolmogorov operator. Because a numerical integrator is adopted to solve the original 
SDE, the resulting Kolmogorov operator, denoted as $\tilde{P}_h^t$, approximates $e^{h \tilde{\mathcal{L}}_t}$.
To characterise the accuracy of a numerical integrator, we use the following definition.

\begin{definition}\label{def:k-order-integrator}
	A numerical integrator is said to be a $K$th-order local integrator if for any smooth and bounded function $f$,
	the corresponding Kolmogorov operator $\tilde{P}^t_h f(\xv)$ from the $t$-th minibatch with stepsize $h$ satisfies the following relation:
	\begin{align}
	\tilde{P}^t_h f(\xv) = e^{h\tilde{\Lcal}_t} f(\xv) + O(h^{K+1})~.
	\end{align}
\end{definition}

We follow~\cite{chen2015integrator},
and state in Lemma~\ref{lem:euler} that a Euler integrator satisfies Definition~\ref{def:k-order-integrator} with $K = 1$ when used in  mSGNHT. Detailed proofs are provided in the Appendix.
\begin{lemma}\label{lem:euler}
	The Euler integrator used in mSGNHT-E is a 1st-order local integrator, {\it i.e.},
	\begin{align}
	\tilde{P}_{h}^{l} = e^{h\tilde{\mathcal{L}}_l} + O(h^2)~.
	\end{align}
\end{lemma}

Using the Baker--Campbell--Hausdorff formula for commutators \cite{Rossmann02}, the SSI can be shown to be 
a 2nd-order integrator in mSGNHT, stated in Lemma~\ref{lem:ssi}.
\begin{lemma}\label{lem:ssi}
	The symmetric splitting integrator used in mSGNHT-S is a 2nd-order local integrator, {\it i.e.},
	\begin{align}
	\tilde{P}_{h}^{l} = e^{h\tilde{\mathcal{L}}_l} + O(h^3)~.
	\end{align}
\end{lemma}

The authors of \cite{chen2015integrator} formalize the role of numerical integrators in terms of posterior {\em bias} and
{\em mean square error} (MSE). Specifically, for a testing function $\phi(x)$,  they study the difference between the posterior average $\bar{\phi} \triangleq \int \phi(x) \rho(x) \mathrm{d}x$ and the finite-time sample average 
$\hat{\phi} \triangleq \frac{1}{T}\sum_{t=1}^T \phi(x_t)$, where $\rho(x)$ denotes the true posterior of
a Bayesian model, and $\{x_t\}$ denotes samples from a SG-MCMC algorithm. To study the role of the SSI
applied in mSGNHT, we simplify the notation and conclude their results in the following lemma.

\begin{lemma}[Roles of numerical integrators] \label{lem:numerical_integrator}\label{lem:role_NI}
	Under certain assumptions, the {\em Bias} and {\em MSE} of a SG-MCMC algorithm with stepsize $h$ 
	and a $K$th-order integrator are:
	\begin{align*}
		\mbox{{\em Bias: }}& \left|\mathbb{E}\hat{\phi} - \bar{\phi}\right| = \mathcal{B}_{\text{bias}} + O(h^K) \\
		\mbox{{\em MSE: }}& \mathbb{E}\left(\hat{\phi} - \hat{\phi}\right)^2 = \mathcal{B}_{\text{mse}} + O(h^{2K})~,
	\end{align*}
	where $\mathcal{B}_{\text{bias}}$ and $\mathcal{B}_{\text{mse}}$ are functions depending on $(h, T)$ but independent of $K$.
\end{lemma}

Based on Lemma~\ref{lem:numerical_integrator} and~\cite{chen2015integrator}, we summarize the properties of mSGNHT-S in the following remarks. The detailed are provided in Appendix.
% In addition, the SSI also endows a faster convergence rate than the Euler integrator:
% Finally, mSGNHT\_S also dominates mSGNHT-E in term of convergence accuracy.

\begin{remark}[Robustness] \label{rem:robustness} 
When applying Lemma~\ref{lem:numerical_integrator} to mSGNHT, the bias and MSE of mSGNHT-S is 
bounded as: $ \Bcal_{\text{bias}} +  O(h^2)$ and $ \Bcal_{\text{mse}} +  O(h^4)$, compared to 
$\Bcal_{\text{bias}} +  O(h)$ and $\Bcal_{\text{mse}} +  O(h^2)$ for the mSGNHT-E, respectively.
This indicates that mSGNHT-S is more robust to the stepsizes than mSGNHT-E.
\end{remark}

\begin{remark}[Convergence Rate] \label{rem:converge} The higher order a numerical integrator is, the faster its optimal convergence rate is. 
Convergence rates in term of {\em bias} for mSGNHT-S and mSGNHT-E
are $T^{-2/3}$ and $T^{-1/2}$, respectively, indicating mSGNHT-S converges faster.
\end{remark}

\begin{remark}[Measure Accuracy] \label{rem:accuracy} 
In the limit of infinite time ($T \rightarrow \infty$), the terms $\mathcal{B}_{\text{bias}}$ 
and $\mathcal{B}_{\text{mse}}$ in Lemma~\ref{lem:role_NI} vanish, leaving only the $O(h^K)$ terms.
This indicates mSGNHT-S is an order of magnitude more accurate than mSGNHT-E.
\end{remark}

\subsection{Advantages of mSGNHT-S for Deep Learning}

Compared to optimization-based methods~\cite{martens2010deep}, the mSGNHT-S is able to more fully explore the parameter space. Therefore, it is less sensitive to initializations, a nontrivial issue in optimization~\cite{sutskever2013importance}. Second, the mSGNHT is related to stochastic gradient descent (SGD) with momentum in optimization~\cite{ChenFG:ICML14,DingFBCSN:NIPS14}, with the additional advantage that the momentum is updated element-wise to automatically adapt stepsizes. Additionally, \cite{sutskever2013importance} shows that momentum-accelerated  SGD is capable of accelerating along directions of low-curvature in the parameter space, leading to faster convergence speed. As a result, mSGNHT is more favorable than other momentum-free SG-MCMC algorithms, such as the ``vanilla'' SGLD \cite{welling2011bayesian}.

Specifically for mSGNHT, we know from previous analysis that $(\RN{1})$ mSGNHT-S is less sensitive to stepsize as shown in Remark~\ref{rem:robustness}, and thus can tolerate gradients of various magnitudes.
This provides a potential solution to mitigate the {\em vanishing/exploding gradients} problem \cite{williams1986learning}. Our empirical results on deep neural networks verifies this in Section 5.2.
$(\RN{2})$  Convergence speed is a critical criteria for learning, and mSGNHT-S clearly outperforms mSGNHT-E  in this regard as discussed in Remark~\ref{rem:converge}. $(\RN{3})$ mSGNHT-S converges to a solution one order magnitude more accurate than mSGNHT-E, as discussed in Remark~\ref{rem:accuracy}, and it is more accurate in estimating model parameters. The number of parameters in large-scale models may be significant, and small numerical error in individual parameters can accumulate, causing noticeable inefficiency. For these reasons, we advocate mSGNHT-S for training large deep models.

\vspace{-4mm}
\section{Related Work}

One direction for scalable Bayesian learning of deep models is stochastic variational inference. For deep models with stochastic hidden units, learning has been implemented using variational methods; when the latent variables are continuous, Stochastic Gradient Variational Bayes (SGVB)~\cite{kingma2013auto} has been employed, while models with discrete latent variables have been trained via Neural Variational Inference and Learning (NVIL)~\cite{mnih2014neural}. For deep models with deterministic hidden units, recent studies have shown that imposing uncertainty on global parameters helps prevent overfitting, yielding significant improvment on model performances. Representative works of this type include Bayes by Backprop (BBB)~\cite{blundell2015weight} and Probabilistic Backpropagation (PBP)~\cite{hernandez2015probabilistic}, which approximate posteriors of network weights as a product of univariate Gaussian distributions.
% is to  estimate second-order information, and rescale parameters so that the loss function has similar curvature along all directions. This strategy has shown improved performance,  including methods

Another direction for Bayesian deep learning is SG-MCMC, the line of work followed by this paper. These methods do not have to assume a simplifying form for the posterior, as in variational methods. SG-MCMC algorithms are closely related to optimization methods (\eg SGD). The difference with respect to optimization methods is the injection of Gaussian noise in the parameter update, allowing better exploration of parameter space when learning. Works of this type include the SGLD~\cite{welling2011bayesian}, SGHMC~\cite{ChenFG:ICML14}, SGNHT \cite{DingFBCSN:NIPS14}, and mSGNHT~\cite{gan2015scalable}.
% {\color{red}Training deep models is complicated by the fact that distribution of each layer’s inputs changes during training, as the parameters of the previous layers change.} This would probably lead to 
% The pathological curvature in the parameter space usually slows down training by requiring lower learning rates, and makes it hard to train models with saturating nonlinearities~\cite{ioffe2015batch}.
% Fortunately, 
It has been shown in~\cite{sutskever2013importance} that carefully tuned momentum methods suffice for dealing with curvature issues in deep network training. mSGNHT belongs to the class of momentum-accelerated SG-MCMC algorithms. In terms of numerical integrators, recently work for HMC include~\cite{chaoexponential,shahbaba2014split}.
For our SDE setting, \cite{leimkuhler2015adaptive} proposes a symmetric splitting scheme for the SGNHT with a specific stochastic gradient noise, which is different from our setting of mSGNHT for deep models.  Recently, \cite{chen2015integrator} provides a theoretical foundation for rigorous study of numerical integrators for SG-MCMC. Our work is complementary, providing implementation guidance for the numerical integrator for other SG-MCMC, and investigating its roles in practical applications.

\vspace{-0mm}

\begin{figure*} \centering
	\hspace{-5mm}
	\begin{tabular}{ c c c } %\hline
		\begin{minipage}{5.0cm}
			\includegraphics[width=5.0cm]{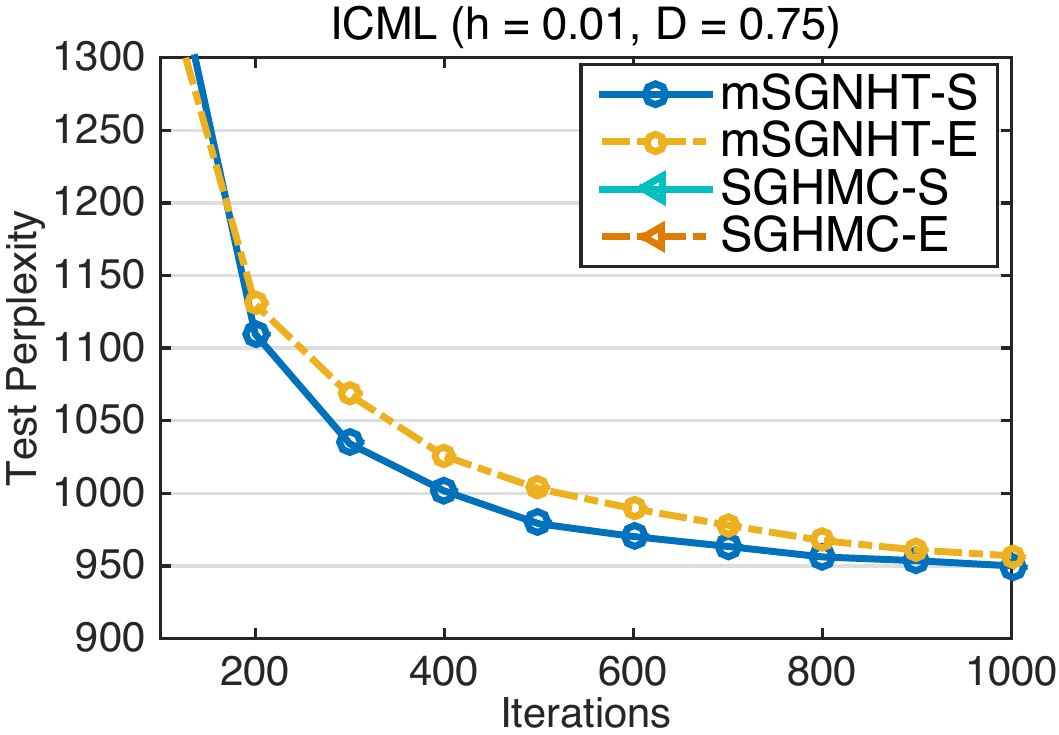} 
		\end{minipage}
		\hspace{-0mm}
		\begin{minipage}{5.0cm}
			\includegraphics[width=5.0cm]{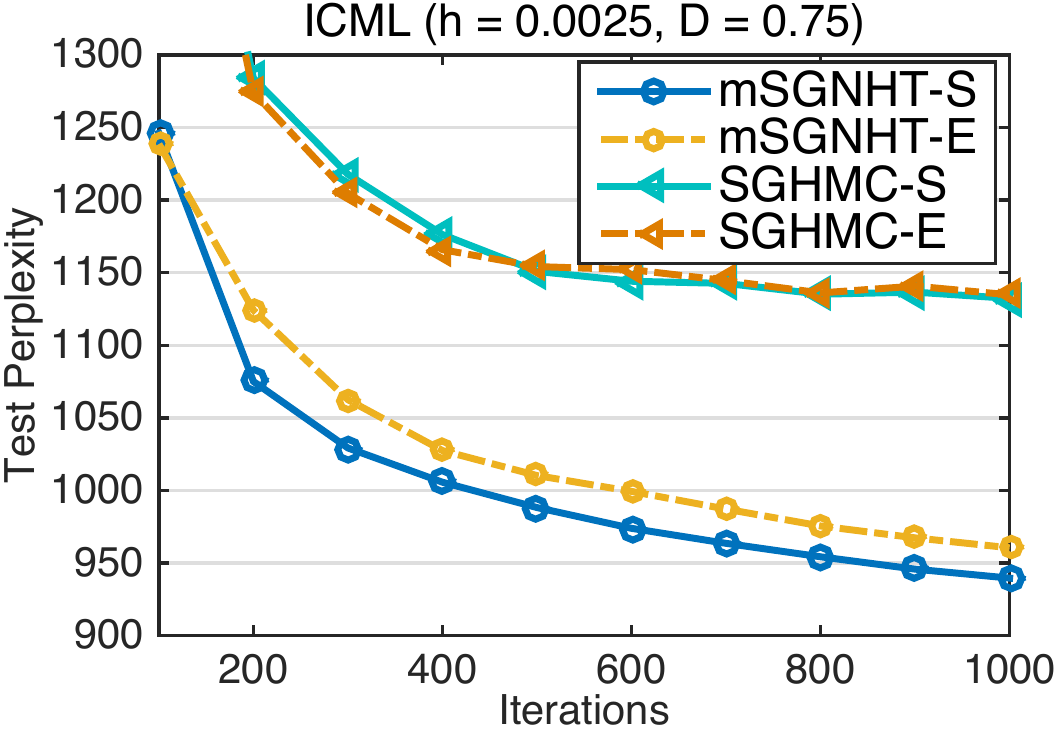} 
		\end{minipage} 
		\hspace{-0mm}
		\begin{minipage}{5.0cm}
			\includegraphics[width=5.0cm]{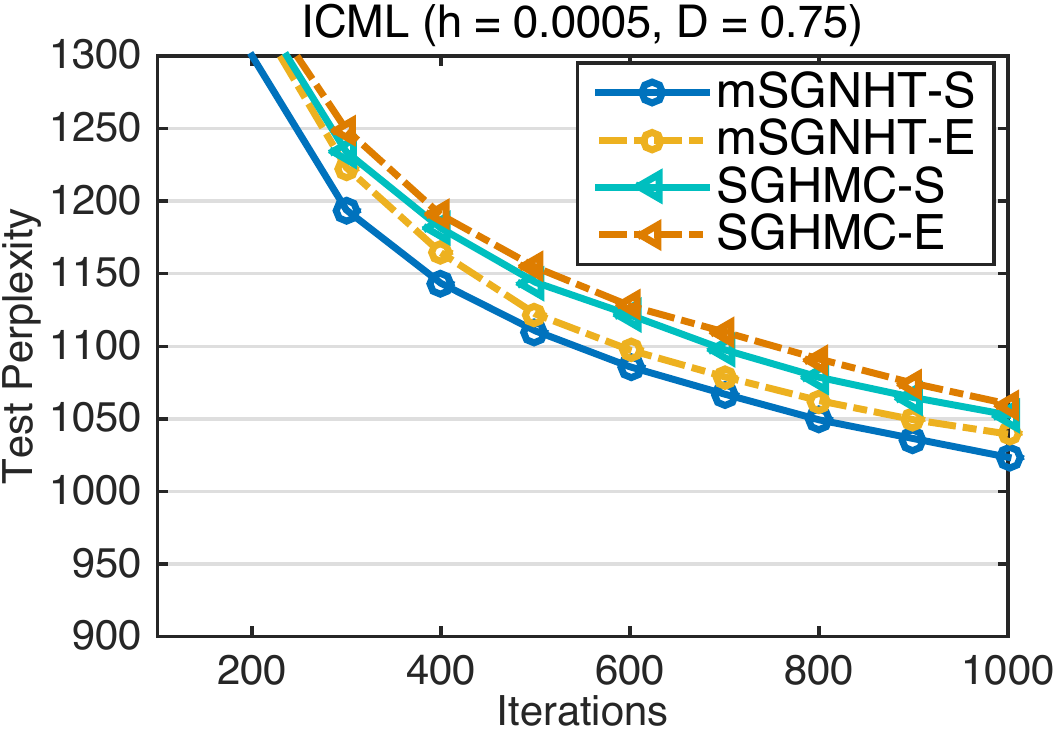} 
		\end{minipage}	
	\end{tabular}
	\vspace{-3mm}
	\caption{Learning curves of LDA on $\mathtt{ICML}$ dataset for different stepsize $h$.}		
	\label{fig:lda-icml}
	\vspace{-4mm}
\end{figure*}

\newpage 
\section{Experiments}

\vspace{-0mm}
\subsection{Canonical Models}
We consider two representative Bayesian models to demonstrate that mSGNHT-S improves general posterior sampling: Latent Dirichlet Allocation (LDA)~\cite{blei2003latent} for latent variable models,  and logistic regression.

\vspace{-0mm}
\subsubsection{Latent Dirichlet Allocation} 
We first evaluate our method on the $\mathtt{ICML}$ dataset~\cite{chen2015differential} using LDA. This dataset contains $765$ documents from the abstracts of ICML proceedings from 2007 to 2011. After removing stopwords, we obtained a vocabulary size of $1918$ and total words of $44140$. We used $80\%$ of the documents (selected at random) for training and
the remaining $20$\% for testing. Similar to~\cite{patterson2013stochastic}, we used the semi-collapsed LDA whose posterior is provided in the Appendix. 
Following~\cite{DingFBCSN:NIPS14}, a Gaussian prior $\Ncal(0.1, 1)$ is used for the reparametrized parameter.
The Dirichlet prior parameter for topic distribution for each document is set to $0.1$. The number of topics is set to $30$. We use {\em perplexity}~\cite{blei2003latent} to measure the quality of algorithms.

To show the robustness of mSGNHT-S to stochastic gradient noise, we chose minibatch of size 5, and
$D$ in mSGNHT-S is fixed as $0.75$. We test a wide range of  values for step size $h$. Generally, larger $h$ imposes larger gradient-estimation error and numerical error. Learning curves of the test perplexity for $h=10^{-2}, 2.5\!\times\!10^{-3}, 5\!\times\!10^{-4} $ are shown in Fig.~\ref{fig:lda-icml}. We observe that the proposed SSI is consistently better than the Euler integrator. Furthermore, mSGNHT is shown to significantly outperform the SGHMC when $h$ is large. We note that the gap between SSI and the Euler integrator is larger for mSGNHT's than SGHMC's, indicating the importance of numerical integrators in higher dimensional systems. 

The best performances for each method are shown in Table~\ref{table:lda}. Note Gibbs sampling typically obtains the best perplexity because it uses the full dataset for each update. However, it is not scalable for large datasets. In our noisy gradient setup, we see that mSGNHT-S provides the lowest perplexity among the SG-MCMC methods, including a stochastic sampling method for simplex-structured distributions, Stochastic Gradient Riemannian Langevin Dyanmics (SGRLD)~\cite{patterson2013stochastic}.

\begin{table} [t!]
	\begin{minipage}{.47\linewidth}
		%\vskip -0.3in
		\caption{\small LDA on $\mathtt{ICML}$.} \label{table:lda}
		\vskip -0.1in
		\centering
		%\raggedleft
		\small
		\begin{sc}
			\begin{adjustbox}{scale=0.9,tabular=lc,center}
				\hline
				% \abovespace \belowspace
				{\bf Method} & {\bf Test Perplexity $\downarrow$} \\
				\hline 
				mSGNHT-S     &  {\bf 939.67}\\
				mSGNHT-E				     &  960.56\\	
				SGHMC-S     &  1004.73\\   % 1.00473 (SGHMCS_h_0.0005_B_0.1)
				SGHMC-E				     &  1017.51\\			% 1.01751(SGHMCE_h_0.0005_B_0.05)		  
				\hline			
				SGRLD             &  1154.68\\		 	
				Gibbs			    		       &  907.84\\		 	
				\hline 
			\end{adjustbox}
		\end{sc}
	\end{minipage}
	\hspace{0.2cm}
	\begin{minipage}{.47\linewidth}
		%\vskip -0.3in
		\caption{LR on $\mathtt{a9a}$.} \label{table:a9a}
		\vskip -0.1in
		\centering
		%\raggedleft
		\small
		\begin{sc}		
			\begin{adjustbox}{scale=0.9,tabular=lc,center}
				\hline   
				{\bf Method} &    {\bf Test Accuracy $\uparrow$}    \\
				\hline
				mSGNHT-S &  {\bf 84.95\%} \\ 
				mSGNHT-E &   84.72\%  \\
				SGHMC-S & 84.56\%  \\
				SGHMC-E &  84.51\%	 \\
				\hline
				DSVI       & 84.80\% \\ 	 	
				HFSGVI   &  84.84\%\\		 			
				\hline
			\end{adjustbox}
		\end{sc}
	\end{minipage}
	\vspace{-5mm}
\end{table}

\subsubsection{Logistic Regression}
We examine logistic regression (LR) on the $\mathtt{a9a}$ dataset\footnote{http://www.csie.ntu.edu.tw/ ̃cjlin/libsvmtools/datasets/binary.html}. The training and testing data consist of $32561$ and $16281$ data points, respectively, with parameter dimension $123$.
The minibatch size is set to $10$, and  the Gaussian prior on the parameters is $\Ncal(0, 10)$.
A thinning interval of $50$ is used, with burn-in $300$, and $3\!\times\!10^3$ total iterations. Similar to the experiments for LDA, we test a large range of $h$. We find that mSGNHT-S gives stable performances across varying $h$ on the regression model. Test accuracies are compared in Table~\ref{table:a9a}, from which mSGNHT-S also outperforms the recent doubly stochastic variational Bayes (SDVI)~\cite{titsias2014doubly}, and a higher-order variational autoencoder method (HFSGVI)~\cite{fan2015secondVAE}. More details are provided in the Appendix.

\subsection{Deep Models}
To illustrate the advantages of the proposed algorithm for deep learning, two deep models are considered. Specifically, for the case of deterministic hidden layers, we consider deep Feedforward (Convolutional) Neural Networks (FNN), and for stochastic latent layers, we consider Deep Poisson Factor Analysis (DPFA) \cite{gan2015learning}.
%Bayesian Feedforward Neural Network (BFNN) extends BLR with deterministic layers, \ie composition of nonlinearities; With the connections between LDA and Poission Factor Analysis (PFA)~\cite{zhou2012beta} shown in~\cite{zhou2015negative}, Deep Poission Factor Analysis (DPFA) can be considered extending LDA with stochastic latent layers. 
% Deep generative models.

\subsubsection{Deep Neural Networks} We evaluate FNN on the MNIST dataset for classification. The data contains $60000$ training examples and $10000$ testing examples, each being an $28\times 28$ image of handwritten digit.  A $L$-layer FNN is equivalent to compositing $L$ times of a nonlinear function $g_{\thetav_{\ell}}$, \eg the sigmoid function used in the logistic regression model. At the top layer, a softmax function is used for multi-class classification, specifically
%
%\begin{align*}
$$
P(y | \xv) \propto \text{softmax} \big ( g_{\thetav_L}  \circ  \cdots  \circ  g_{\thetav_0} (\xv  )   \big )~,
$$
where $\circ$ denotes function composition. 
For each data $\{\xv, y\}$, $\xv \in \R^{784}$ is the raw image, $y$ is the label.
A Gaussian prior is placed
on model parameters $\thetav = \{ \thetav_0, \dots, \thetav_L\} \propto \Ncal(0, \sigma^2 \Imat)$ with $\sigma^2  =1$ in our experiment.

\begin{figure}[t!] \tiny  \centering
	\begin{tabular}{ c  c  } 
		\hspace{-4mm}
		\begin{minipage}{3.6cm}
			\includegraphics[width=3.6cm]{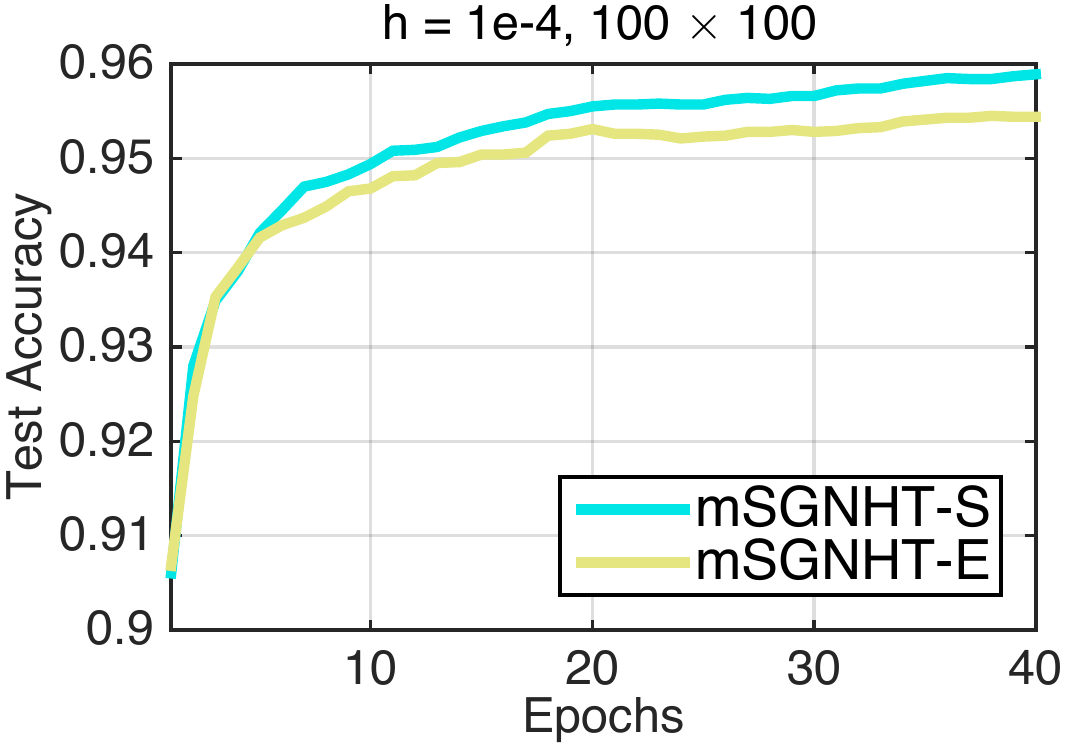} 
		\end{minipage}   &
		\hspace{-2mm}
		\begin{minipage}{3.6cm}
			\includegraphics[width=3.6cm]{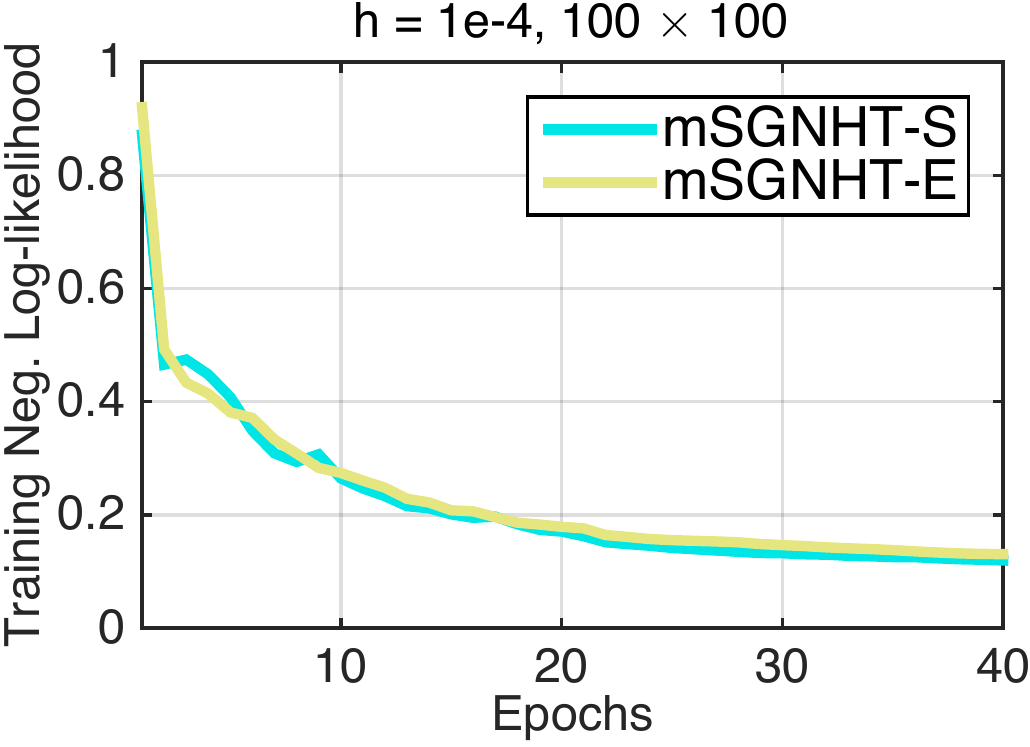} 
		\end{minipage}  \\		
		\hspace{-5mm}
		\begin{minipage}{3.6cm}
			\includegraphics[width=3.6cm]{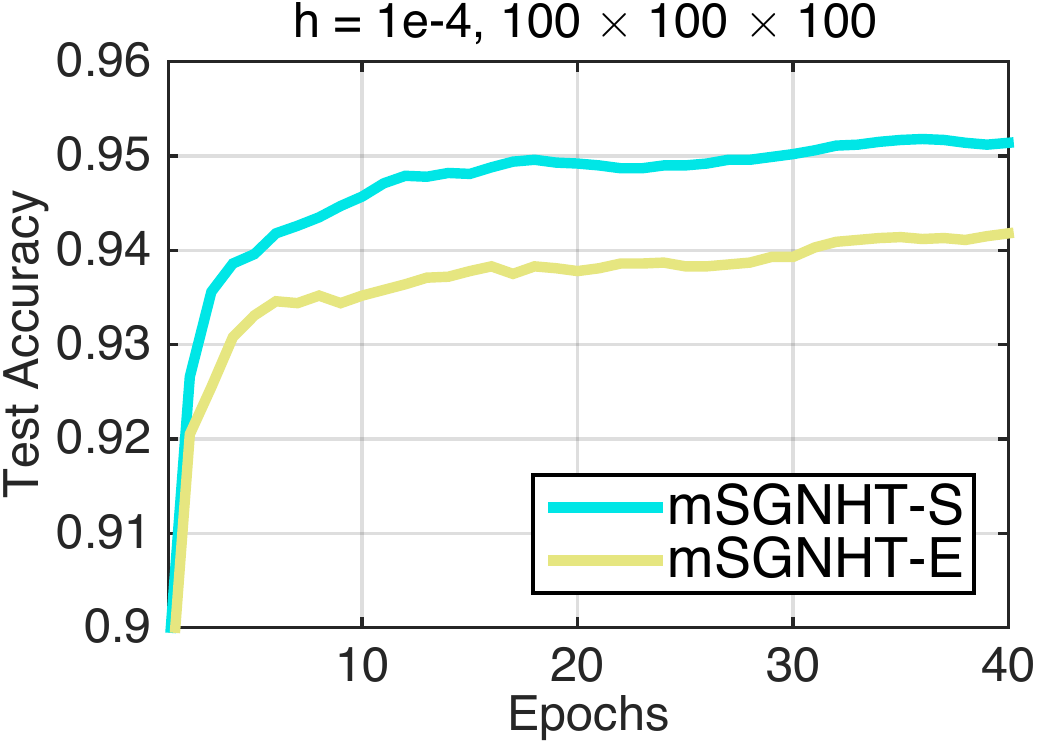} 
		\end{minipage}  &
		\hspace{-2mm}
		\begin{minipage}{3.6cm}
			\includegraphics[width=3.6cm]{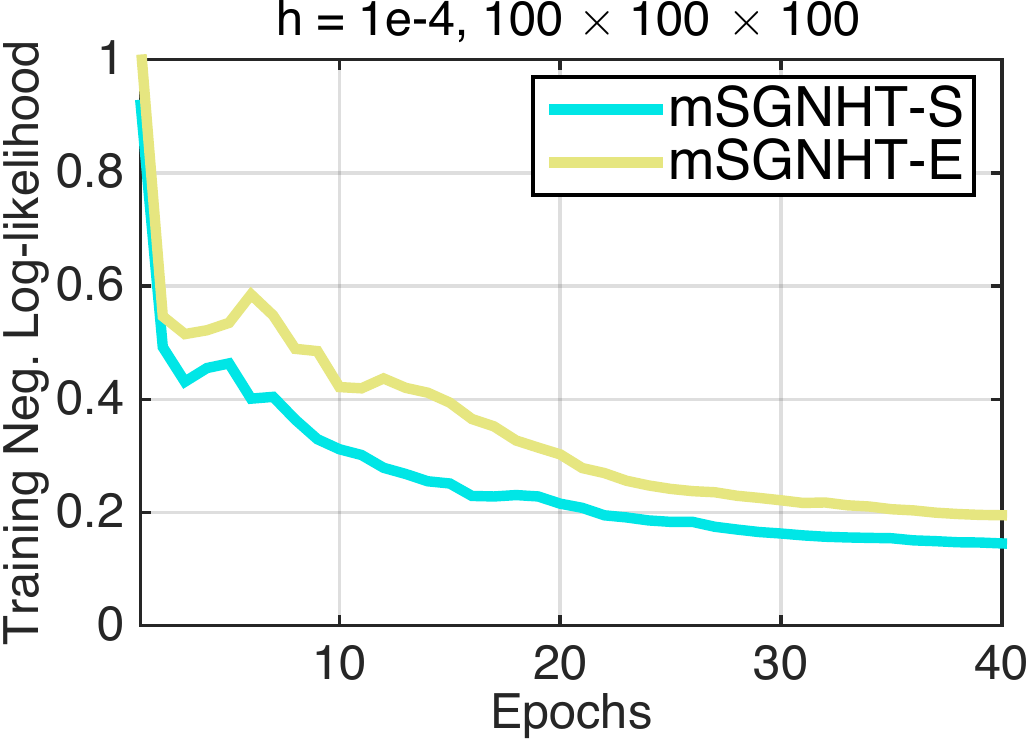} 
		\end{minipage}   \\		
		\hspace{-5mm}
		\begin{minipage}{3.6cm}
			\includegraphics[width=3.6cm]{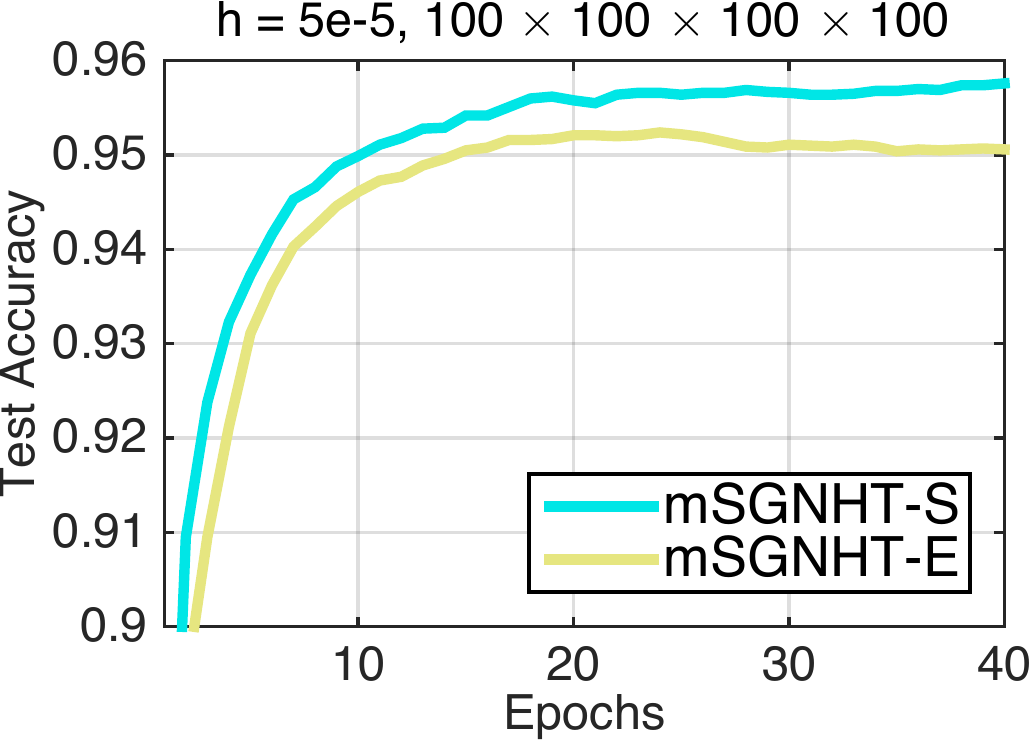} 
		\end{minipage} &
		\hspace{-2mm}
		\begin{minipage}{3.6cm}\vspace{0mm}
			\includegraphics[width=3.6cm]{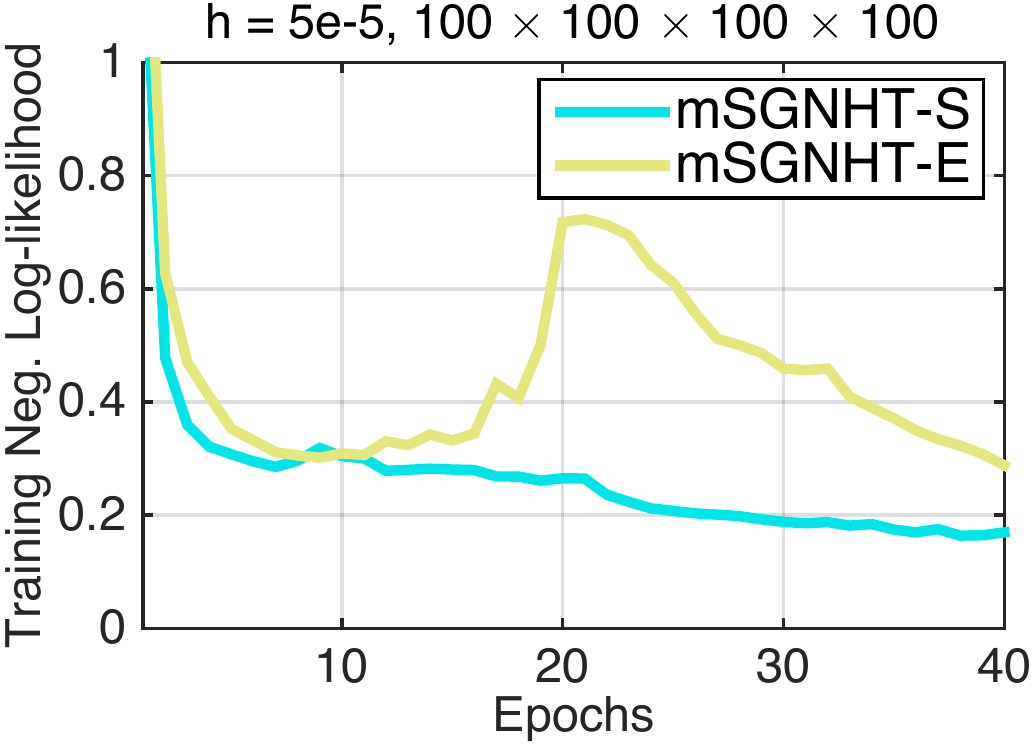} 
		\end{minipage} \\
	\end{tabular}
	\vspace{-2mm}
	\caption{Learning curves of FNN with different depth on MNIST dataset.}			
	\label{fig:bnn-mnist}
	\vspace{-5mm}
\end{figure}

We use the Rectified Linear Unit (ReLU)~\cite{glorot2011deep} as $g_{\thetav_{\ell}}$ in each layer. The number of hidden units for each layer is $100$, $D$ is set to $5$, 
stepsize $h$ is set to $10^{-4}$, $40$ epochs are used. To reduce bias~\cite{chen2015integrator}, $h$ is decreased by half at epoch $20$. We test the FNNs with depth $\{2, 3, 4\}$, respectively. Fig.~\ref{fig:bnn-mnist} displays learning curves on testing accuracy and training negative log-likelihood. It can be seen that mSGNHT-S consistently converges faster, better than mSGNHT-E for both training and testing. The gaps between mSGNHT-S and mSGNHT-E becomes larger in deeper models. Notably, in the 4-layer FNN, mSGNHT-E failed when $h=10^{-4}$, while mSGNHT-S worked well. We therefore plot the results for $h=5\!\times\!10^{-5}$. 
From the training plot, mSGNHT-E is failing as learning progresses. It starts to work, only because the stepsize  is decreased by half.
% ({\color{red}this is weird, typically you can just stop here plotting instead of decreasing the step size}). 
This confirms that mSGNHT-S is robust to stepsizes, thus is able to mitigate the vanishing/exploding gradient problem in deep models. In addition, we also consider $g_{\thetav_{\ell}}$ as the sigmoid activation function, and the case of convolutional neural networks, empirical results are consistent with the ReLU case. More results are in the Appendix.
%
%\vspace{-3mm}

\vspace{-4mm}
\subsubsection{Deep Poisson Factor Analysis}
DPFA~\cite{gan2015scalable} is a recently proposed framework for deep topic modeling, where interactions between topics are inferred through a deep latent binary hierarchy. We adopt the deep sigmoid belief networks (DSBN)~\cite{gan2015learning} as the deep architecture in the experiment. The generative process is
\vspace{-1mm}
%
%\begin{align*}
$$
\Wmat \sim \Pois(\Phimat (\Psimat   \odot \hv^{(1)} )), \quad \hv^{(1)}  \sim \text{DSBN}(\hv^{ \{(2), \cdots, (L) \} }),
$$
% \vspace{-2mm}
%\end{align*}
%
where $\Wmat \in \mathbb{Z}_+^{V\times J}$ is the observed word count matrix for $J$ documents and vocabulary 
size $V$.
$\Phimat$ is the word-topic matrix, column $k$, $\phiv_k \in \bigtriangleup_V$, encodes the relative importance of each word in topic $k$, with $\bigtriangleup_V$ representing the $V$-dimensional simplex.
$\Psimat$ is the topic-document matrix, each of its column $\psi_j$ contains relative topic intensities specific to document $j$. The latent binary feature matrix $\hv^{(1)}$ indicates the usage of topic. We use mSGNHT to infer the parameters in DSBN, and the {\em Expanded-Natural} reparametrization method to sample from the probabilistic 
simplex~\cite{patterson2013stochastic}.
% The bottom binary layer $\hv$ selects topics for use, while the deep specification in DSBN serves as a flexible prior for revealing topic structure. 
More details for model specification are in the Appendix.

We test the DPFA on a large dataset, \emph{Wikipedia}, from which $10$M randomly 
downloaded documents are used, using scripts provided in \cite{hoffman2010online}. 
We follow the setup in~\cite{gan2015scalable}, where 1K documents are randomly selected for testing and validation, respectively. The vocabulary size is $7702$, and the minibatch size is set to $100$, with one pass of the whole data
in the experiments. We collect $300$ posterior samples 
to calculate test perplexities, with a standard holdout technique. A three-layer DSBN is employed, with dimensions $128$-$64$-$32$  ($128$ topics right above the data layer).
Step sizes are chosen as $10^{-4}$ and $10^{-5}$, and parameter $D = 40$. 

% One important protocal to evaluate topic modeling is the test perplexity, which reduces to evalate different training methods,  provided the same DPFA model. 

\begin{figure}[t!]
	\centering	
	\begin{minipage}{8.5cm}\centering	
		\includegraphics[width=8.5cm]{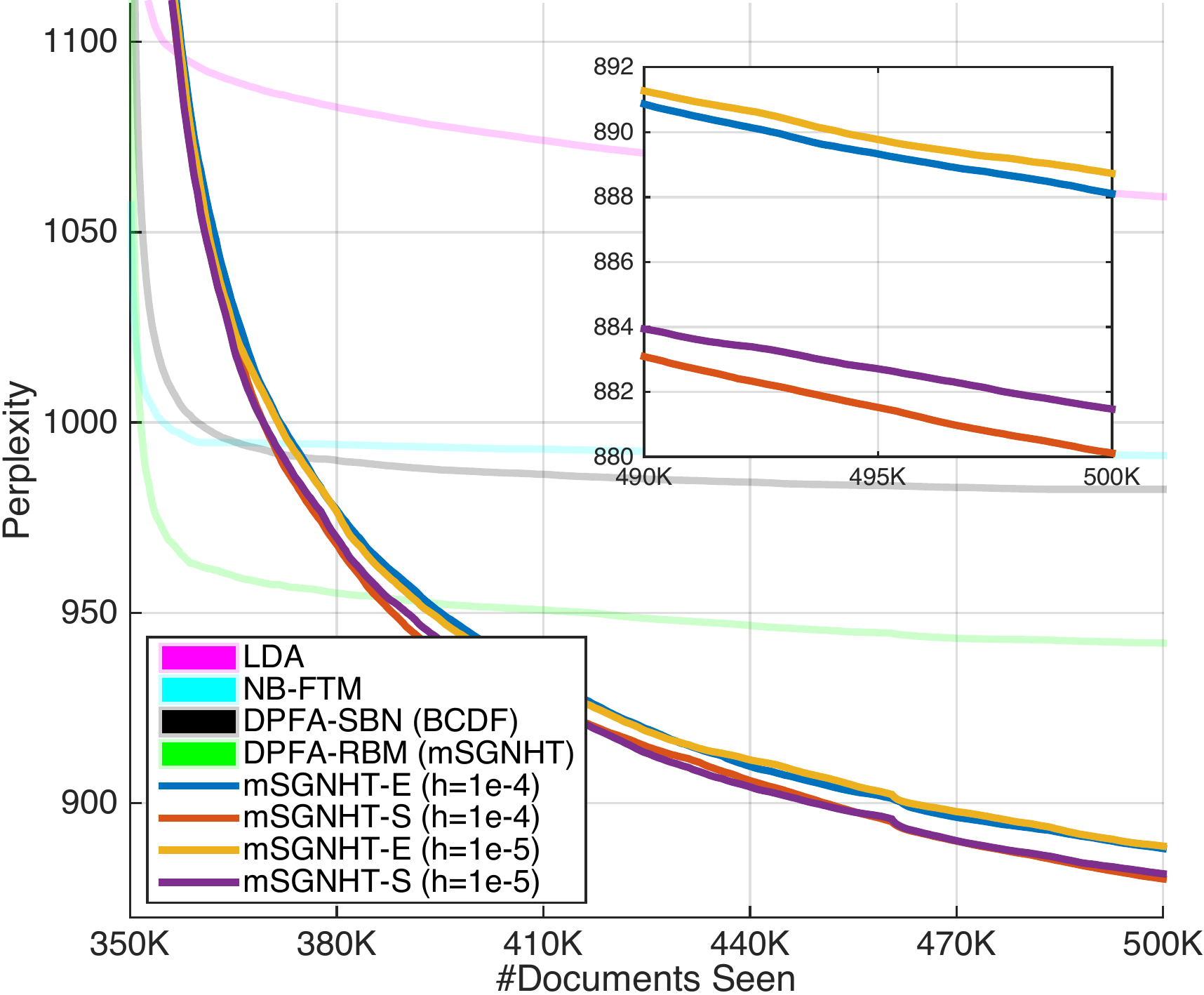} 
	\end{minipage}  
	\vspace{-2mm}
	\caption{Perplexity on Wikipedia dataset.}\label{fig:wikipedia}
	\vspace{-4mm}
\end{figure}

The results are shown in Fig.~\ref{fig:wikipedia}, displaying the predictive perplexities on a held-out test set as a function of training documents seen. Clearly, mSGNHT-S converges faster than mSGNHT-E at both chosen stepsizes. A magnified plot is shown at the top-right corner of the figure as well, displaying perplexities for the last $10$K documents.

mSGNHT-S outperforms other recent state-of-the-art methods (shown in semi-transparent plots). Specifically, we compare to, DPFA-SBN trained with Bayesian conditional density filtering (BCDF)~\cite{guhaniyogi2014bayesian}, DPFA with restricted Boltzmann machines (RBM)~\cite{hinton2002training} trained with mSGNHT-E, and Negative Binomial Focused Topic Model (NB-FTM)~\cite{zhou2015negative} trained with BCDF. The shallow model LDA trained with BCDF is reported as the baseline.

%  mSGNHT-S outperforms other methods, including DPFA-SBN trained with Bayesian conditional density filtering (BCDF) algorithm~\cite{guhaniyogi2014bayesian}, DPFA with restricted Boltzmann machine (RBM)~\cite{hinton2002training} trained with mSGNHT-E, and Negative Binomial Focused Topic Model (NB-FTM)~\cite{zhou2015negative} trained with BCDF.

\vspace{-2mm}
\section{Conclusion}
A 2nd-order symmetric splitting integrator is proposed to solve the SDE within mSGNHT.
This method is shown to be more accurate than the conventional Euler integrator, leading to higher robusness, faster convergence, and more accurate posterior samples. We apply the integrator on mSGNHT for four representative models, including 
latent Dirichlet allocation, logistic regression, deep neural networks, and deep Poisson factor analysis. Extensive experiments demonstrate that the proposed scheme improves large-scale sampling in terms of convergence speed and accuracy, particularly for deep models.
%A 2nd-order symmetric splitting method is proposed to more accurately solve the SDE of the mSGNHT.  
%This is crutial to mSGNHT, because there are more auxiliary variables involved in its continuous-time diffusion. Compared with the conventional Euler integrator,  we show that the proposed scheme is more accurate, robust, and converges faster. Importantly, these properties are desirable for Bayesian deep models. We apply our technique to four representative Bayesian models: LDA and DPFA (latent variable model and its deep extension with stochastic layers), BLR and BFNN (regression model and its deep extension with deterministic layers).  Extensive experiments demonstrate that the proposed scheme improve general Bayesian posterior sampling, particularly for deep models.

\newpage
\paragraph{Acknowledgements} This research was supported in part by ARO, DARPA, DOE, NGA, ONR and NSF.

\bibliographystyle{aaai}	
\bibliography{subtex/references.bib}

\newpage

\appendix
\input{sgnht_supp}

\end{document}

%% file: subtex/mlVecMat.tex
\usepackage{hyperref}
\usepackage{url}

\usepackage{amsmath}
\usepackage{amsfonts}
\usepackage{amssymb}
\usepackage{wrapfig}
\usepackage{bm}
\usepackage{subcaption}
\usepackage{psfrag}
\usepackage{tikz}
\usepackage{epstopdf}

\usepackage{color}
\usepackage{tikz}
\usetikzlibrary{arrows,shapes,snakes,automata,backgrounds,fit,petri}
\usepackage{adjustbox}
\usepackage{gensymb }
\usepackage{algorithm}
\usepackage{algorithmic}

\newcommand{\ie}[0]{\emph{i.e., }}

\newcommand{\eg}[0]{\emph{e.g., }}

\newcommand{\beq}{\vspace{0mm}\begin{equation}}
\newcommand{\eeq}{\vspace{0mm}\end{equation}}
\newcommand{\beqs}{\vspace{0mm}\begin{eqnarray}}
\newcommand{\eeqs}{\vspace{0mm}\end{eqnarray}}
\newcommand{\barr}{\begin{array}}
\newcommand{\earr}{\end{array}}

\newcommand{\Amat}[0]{{{\bf A}}}
\newcommand{\Bmat}{{\bf B}}

\newcommand{\Imat}{{\bf I}}

\newcommand{\Wmat}[0]{{{\bf W}}}
\newcommand{\Xmat}[0]{{{\bf X}}}

\newcommand{\av}[0]{{\boldsymbol{a}}}
\newcommand{\bv}[0]{{\boldsymbol{b}}}
\newcommand{\cv}[0]{{\boldsymbol{c}}}

\newcommand{\hv}[0]{{\boldsymbol{h}}}

\newcommand{\pv}[0]{{\boldsymbol{p}}}

\newcommand{\wv}{\boldsymbol{w}}
\newcommand{\xv}{\boldsymbol{x}}

\newcommand{\zv}{\boldsymbol{z}}

\newcommand{\Thetamat}{\boldsymbol{\Theta}}

\newcommand{\Phimat}{\boldsymbol{\Phi}}
\newcommand{\Psimat}{\boldsymbol{\Psi}}

\newcommand{\zetav}{\boldsymbol{\zeta}}

\newcommand{\thetav}{\boldsymbol{\theta}}

\newcommand{\xiv}[0]{{\boldsymbol{\xi}}}

\newcommand{\phiv}{\boldsymbol{\phi}}

\newcommand{\psiv}{\boldsymbol{\psi}}

\newcommand{\R}{\mathbb{R}}

\newcommand{\Z}{\mathbb{Z}}

\newcommand{\Pois}{\text{Pois}}
\newcommand{\Gal}{\text{Gamma}}
\newcommand{\Dir}{\text{Dir}}

\newcommand{\Lcal}{\mathcal{L}}
\newcommand{\Ncal}{\mathcal{N}}
\newcommand{\Bcal}{\mathcal{B}}

\newcommand{\RN}[1]{%
	\textup{\lowercase\expandafter{\it \romannumeral#1}}%
}

\newtheorem{theorem}{Theorem} % [section]
\newtheorem{lemma}[theorem]{Lemma}

\newtheorem{remark}{Remark}
\newtheorem{definition}{Definition}

\ifx\assumption\undefined
\newtheorem{assumption}{Assumption}
\fi

\newenvironment{proof}[1][Proof]{\begin{trivlist}
\item[\hskip \labelsep {\bfseries #1}]}{\end{trivlist}}

\newcommand{\qed}{\nobreak \ifvmode \relax \else
      \ifdim\lastskip<1.5em \hskip-\lastskip
      \hskip1.5em plus0em minus0.5em \fi \nobreak
      \vrule height0.75em width0.5em depth0.25em\fi}

%% file: sgnht_supp.tex
% \def\year{2015}
%File: formatting-instruction.tex
% \documentclass[letterpaper]{article}
%\include{subtex/mlVecMat}
%
%%\newcommand{\RN}[1]{%
%%  \textup{\lowercase\expandafter{\it \romannumeral#1}}%
%%}
%
%\usepackage{aaai}
%\usepackage{times}
%\usepackage{helvet}
%\usepackage{courier}
%
%\usepackage{marvosym}
%\newcommand{\envelope}{\raisebox{-.5pt}{\scalebox{1.05}{\Letter}}\kern1.0pt}

%\frenchspacing
%\setlength{\pdfpagewidth}{8.5in}
%\setlength{\pdfpageheight}{11in}
%\pdfinfo{
%/Title (Insert Your Title Here)
%/Author (Put All Your Authors Here, Separated by Commas)}

% \noindent\makebox[\linewidth]{\rule{\linewidth}{3.5pt}}
\twocolumn[
\begin{center}
	\bf{\LARGE Supplementary Material of \\High-Order Stochastic Gradient Thermostats for\\
		Bayesian Learning of Deep Models\\ \vspace{1em}}
\end{center}
]
%	\noindent\makebox[\linewidth]{\rule{\linewidth}{1pt}}
% 

%\author{Chunyuan Li$^{1}$, Changyou Chen$^{1}$, Kai Fan$^{2}$ and Lawrence Carin$^{1}$\\
%	$^{1}$Department of Electrical and Computer Engineering,  Duke University\\
%	$^{2}$Computational Biology and Bioinformatics, Duke University\\
%	{\footnotesize
%		\href{mailto:chunyuan.li@duke.edu}{  \;\;\;\;\nolinkurl{chunyuan.li@duke.edu,}  } 
%		\href{mailto:cchangyou@gmail.com}{\nolinkurl{cchangyou@gmail.com,}  } 
%		\href{mailto:kai.fan@duke.edu}{\nolinkurl{kai.fan@duke.edu,}  } 
%		\href{mailto:lcarin@duke.edu}{\nolinkurl{lcarin@duke.edu} }
%	}
%}

% \setcounter{secnumdepth}{2}  

%  \begin{document}

%\author{AAAI Press\\
%Association for the Advancement of Artificial Intelligence\\
%2275 East Bayshore Road, Suite 160\\
%Palo Alto, California 94303\\
%}
% \maketitle
% \appendix

\section{The proof of Lemma~1}%\ref{theo:mse_decrease}}

\begin{proof}
In mSGNHT, $\Xmat_{t} = (\thetav_{t}, \pv_{t}, \xiv_{t})$. The update equations with a Euler integrator are:
\begin{align*}
	&\left\{\begin{array}{ll}
	\thetav_{t+1} &= \thetav_t + \pv_t h \\
	\pv_{t+1} &= \pv_t - \nabla_{\thetav}\tilde{U}_t(\thetav_{t+1}) h - \mbox{diag}(\xiv_t) \pv_t h + \sqrt{2D} \zetav_{t+1} \\
	\xiv_{t+1} &= \xiv_t + \left(\pv_{t+1} \odot \pv_{t+1} - 1\right) h
	\end{array}\right.
\end{align*}

Based on the update equations, it is easily seen that the corresponding Kolmogorov operator
$\tilde{P}_{h}^{l}$ for mSGNHT is
\begin{align}\label{eq:sghmc_euler}
	\tilde{P}_{h}^{l} = e^{h\mathcal{L}_1} \circ e^{h\mathcal{L}_2} \circ e^{h\mathcal{L}_3}~,
\end{align}
where $\mathcal{L}_1 \triangleq \left(\pv \odot \pv - 1\right) \cdot \nabla_{\xiv}$,
$\mathcal{L}_2 \triangleq - \mbox{diag}(\xiv) \pv_{t} \cdot \nabla_{\pv} -\nabla_{\thetav}\tilde{U}_l(\thetav) \cdot \nabla_{\pv} + 2DI : \nabla_{\pv} \nabla_{\pv}^T$, and
$\mathcal{L}_3 \triangleq \pv \cdot \nabla_{\thetav}$. Using the BCH formula, we have
\begin{align}\label{eq:bch_euler}
	\tilde{P}_{h}^{l} = e^{h(\mathcal{L}_1 + \mathcal{L}_2 + \mathcal{L}_3)} + O(h^2)~.
\end{align}

On the other hand, the local generator of mSGNHT at the $t$-th iteration can be seen to be:
\begin{align}
	\tilde{\mathcal{L}}_t = \mathcal{L}_1 + \tilde{\mathcal{L}}_2 + \mathcal{L}_3~,
\end{align}
where $\mathcal{L}_1$ and $\mathcal{L}_3$ are defined previously, and 
$\tilde{\mathcal{L}}_2 = - \mbox{diag}(\xiv) \pv \cdot \nabla_{\pv} -\nabla_{\thetav}\tilde{U}_l(\thetav) \cdot \nabla_{\pv} + 2DI : \nabla_{\pv} \nabla_{\pv}^T$. According to the Kolmogorov's backward equation, we have
\begin{align}\label{eq:euler_kov}
	\mathbb{E}[f(\Xmat_{t+1})] = e^{h\tilde{\mathcal{L}}_t} f(\Xmat_{t})~.
\end{align}
Substitute \eqref{eq:euler_kov} into \eqref{eq:bch_euler} and use the fact that $\pv = \pv_t + O(h)$
by Taylor expansion, it is easily seen
\begin{align*}
	\tilde{P}_{h}^{l} = e^{h\tilde{\mathcal{L}}_t + O(h^2)} + O(h^2) = e^{h\tilde{\mathcal{L}}_t} + O(h^2)~.
\end{align*}
This completes the proof.
\end{proof}

\section{The proof of Lemma~2}

\begin{proof}
In symmetric splitting scheme for mSGNHT, according to the splitting in (4) in the main text, the generator 
$\tilde{\Lcal}_t$ is split  into the following sub-generators which can be solved analytically: 
$\tilde{\Lcal}_l = \mathcal{L}_A  + \mathcal{L}_B + \mathcal{L}_{O_l}$, where
\begin{align*}
	&\mathcal{A} \triangleq \mathcal{L}_A = \pv \cdot \nabla_{\thetav} + \left(\pv \odot \pv - 1\right) \cdot \nabla_{\xiv}, \\
	&\mathcal{B} \triangleq \mathcal{L}_B = - \mbox{diag}(\xiv) \pv_{t} \cdot \nabla_{\pv}, \\
	&\mathcal{O}_l \triangleq \mathcal{L}_{O_l} = -\nabla_{\thetav}\tilde{U}_l(\thetav) \cdot \nabla_{\pv} + 2D : \nabla_{\pv} \nabla_{{\pv}}^T~.
\end{align*}

The corresponding Kolmogorov operator $\tilde{P}_{h}^{l}$ for the splitting integrator can be seen to be:
\begin{align*}
\tilde{P}_{h}^{l} \triangleq e^{\frac{h}{2}\mathcal{L}_A} \circ e^{\frac{h}{2}\mathcal{L}_B} \circ e^{h\mathcal{L}_{O_l}} \circ e^{\frac{h}{2}\mathcal{L}_B} \circ e^{\frac{h}{2}\mathcal{L}_A}, 
\end{align*}

In the following we use the Baker--Campbell--Hausdorff (BCH) formula~\cite{Rossmann02} to show that $\tilde{P}_{h}^{l}$ is a 2nd-order integrator. Specifically,
\begin{align}
	e^{\frac{h}{2}\mathcal{A}} e^{\frac{h}{2}\mathcal{B}} &= e^{\frac{h}{2}\mathcal{A} + \frac{h}{2}\mathcal{B} + \frac{h^2}{8}[\mathcal{A}, \mathcal{B}] + \frac{1}{96}\left([\mathcal{A}, [\mathcal{A}, \mathcal{B}]] + [\mathcal{B}, [\mathcal{B}, \mathcal{A}]]\right) + \cdots} \label{eq:bch1}\\
	&= e^{\frac{h}{2}\mathcal{A} + \frac{h}{2}\mathcal{B} + \frac{h^2}{8}[\mathcal{A}, \mathcal{B}]} + O(h^3)~, \label{eq:bch2}
\end{align}
where $[X, Y] \triangleq XY - YX$ is the commutator of $X$ and $Y$, \eqref{eq:bch1} follows from the BCH formula, and \eqref{eq:bch2}
follows by moving high order terms $O(h^3)$ out of the exponential map using Taylor expansion.
Similarly, for the other composition, we have
\begin{align*}
	&e^{h \mathcal{O}_l} e^{\frac{h}{2}\mathcal{A}} e^{\frac{h}{2}\mathcal{B}} = e^{h \mathcal{O}_l} \left(e^{\frac{h}{2}\mathcal{A} + \frac{h}{2}\mathcal{B} + \frac{h^2}{8}[\mathcal{A}, \mathcal{B}]} + O(h^3)\right) \\
	=& e^{h \mathcal{O}_l + \frac{h}{2}\mathcal{A} + \frac{h}{2}\mathcal{B} + \frac{h^2}{8}[\mathcal{A}, \mathcal{B}] + \frac{1}{2}[h \mathcal{O}_l, \frac{h}{2}\mathcal{A} + \frac{h}{2}\mathcal{B} + \frac{h^2}{8}[\mathcal{A}, \mathcal{B}]]} + O(h^3) \\
	=& e^{h \mathcal{O}_l + \frac{h}{2}\mathcal{A} + \frac{h}{2}\mathcal{B} + \frac{h^2}{8}[\mathcal{A}, \mathcal{B}] + \frac{h^2}{4}[\mathcal{O}_l, \mathcal{A}] + \frac{h^2}{4}[\mathcal{O}_l, \mathcal{B}]} + O(h^3) \\
	&e^{\frac{h}{2}\mathcal{A}} e^{h \mathcal{O}_l} e^{\frac{h}{2}\mathcal{A}} e^{\frac{h}{2}\mathcal{B}} \\
	=& e^{\frac{h}{2} \mathcal{A}} \left(e^{h \mathcal{O}_l + \frac{h}{2}\mathcal{A} + \frac{h}{2}\mathcal{B} + \frac{h^2}{8}[\mathcal{A}, \mathcal{B}] + \frac{h^2}{4}[\mathcal{O}_l, \mathcal{A}] + \frac{h^2}{4}[\mathcal{O}_l, \mathcal{B}]} + O(h^3)\right) \\
	=& e^{h \mathcal{O}_l + h\mathcal{A} + \frac{h}{2}\mathcal{B} + \frac{h^2}{4}[\mathcal{A}, \mathcal{B}] + \frac{h^2}{2}[\mathcal{O}_l, \mathcal{B}]} + O(h^3)
\end{align*}
As a result
\begin{align*}
	&\tilde{P}_{h}^{l} \triangleq e^{\frac{h}{2}\mathcal{B}} e^{\frac{h}{2}\mathcal{A}} e^{h \mathcal{Z}} e^{\frac{h}{2}\mathcal{A}} e^{\frac{h}{2}\mathcal{B}}\\
	=& e^{\frac{h}{2} \mathcal{B}} \left(e^{h \mathcal{O}_l + h\mathcal{A} + \frac{h}{2}\mathcal{B} + \frac{h^2}{4}[\mathcal{A}, \mathcal{B}] + \frac{h^2}{2}[\mathcal{O}_l, \mathcal{B}]} + O(h^3)\right) \\
	=& e^{h \mathcal{O}_l + h\mathcal{A} + h\mathcal{B} + \frac{h^2}{4}[\mathcal{A}, \mathcal{B}] + \frac{h^2}{2}[\mathcal{O}_l, \mathcal{B}] + \frac{h^2}{4}[\mathcal{B}, \mathcal{A}] + \frac{h^2}{4}[\mathcal{B}, \mathcal{O}_l] + \frac{h^2}{8}[\mathcal{B}, \mathcal{B}]} + O(h^3)\\
		=& e^{h(\mathcal{B} + \mathcal{A} + \mathcal{O}_l)} + O(h^3) \\
		=& e^{h(\mathcal{L} + \Delta V_l)} + O(h^3) = e^{h\tilde{\mathcal{L}}_l} + O(h^3)~.
\end{align*}
This completes the proof.
\end{proof}

\section{More details on Lemma~3}

Our justification of the symmetric splitting integrator is based on Lemma~3, which is a simplification of the main theorems 
in \cite{chen2015integrator}. For completeness, we give details of their main theorems in this section.

To recap notation, for an It\'{o} diffusion with an invariant measure $\rho(\Xmat)$, the posterior average is defined as:
$\bar{\phi} \triangleq \int_{\mathcal{X}} \phi(\Xmat) \rho(\Xmat) \mathrm{d}x$ 
for some test function $\phi(\Xmat)$ of interest. Given samples 
$(x_{t})_{t=1}^T$ from a SG-MCMC, we use the {\em sample average} $\hat{\phi} \triangleq \frac{1}{T}\sum_{t=1}^T \phi(x_t)$
to approximate $\bar{\phi}$. 
% In the analysis,
%a functional $\psi$ that solves the following \emph{Poisson Equation} is defined:
%\begin{align}\label{eq:PoissonEq1}
%	\mathcal{L} \psi(x_{t}) =  \phi(x_{t}) - \bar{\phi}~.
%\end{align}
%The solution functional $\psi(x_{t})$ characterizes the difference between $\phi(x_{t})$ 
%and the posterior average $\bar{\phi}$ for every $x_{t}$, thus would typically possess a unique 
%solution, which is at least as smooth as $\phi$ under the elliptic or hypoelliptic settings \cite{MattinglyST:JNA10}. 
%For some technical reasons, the following assumptions are required on the solution functional, $\psi$, of 
%the Poisson equation \eqref{eq:PoissonEq1}. 
%
%\begin{assumption}\label{ass:assumption1}
%$\psi$ and its up to 3rd-order derivatives, $\mathcal{D}^k \psi$, are bounded by a
%function $\mathcal{V}$, {\it i.e.}, 
%$\|\mathcal{D}^k \psi\| \leq C_k\mathcal{V}^{p_k}$ for $k=(0, 1, 2, 3)$, $C_k, p_k > 0$. Furthermore, 
%the expectation of $\mathcal{V}$ on $\{\Xmat_{t}\}$ is bounded: $\sup_t \mathbb{E}\mathcal{V}^p(\thetav_{t}) < \infty$, 
%and $\mathcal{V}$ is smooth such that 
%$\sup_{s \in (0, 1)} \mathcal{V}^p\left(s\Xmat + \left(1-s\right)Y\right) \leq C\left(\mathcal{V}^p\left(\Xmat\right) + \mathcal{V}^p\left(Y\right)\right)$, $\forall \Xmat, Y, p \leq \max\{2p_k\}$ for some $C > 0$.
%\end{assumption}

In addition, we define an operator for the $t$-th iteration as:
\begin{align}
	\Delta V_t \triangleq (\nabla_{\theta}\tilde{U}_t - \nabla_{\theta} U) \cdot \nabla_{\pv}~.
\end{align}

Theorem~\ref{theo:bias1} and Theorem~\ref{theo:MSE} summarize the convergence of a SG-MCMC algorithm
with a $K$-th order integrator with respect to the {\em Bias} and {\em MSE}, under certain assumptions. Please refer to \cite{chen2015integrator} for detailed proofs.

\begin{theorem}\label{theo:bias1}
	% Under Assumption~\ref{ass:assumption1}, 
	Let $\left\|\cdot\right\|$ be the operator norm.
	The bias of an SG-MCMC with a $K$th-order integrator at time 
	$\mathcal{T} = hT$ can be bounded as:
	\begin{align*}
	\left|\mathbb{E}\hat{\phi} - \bar{\phi}\right| = O\left(\frac{1}{Th} + \frac{\sum_t \left\|\mathbb{E}\Delta V_t\right\|}{T} + h^K\right)~. 
	\end{align*}
\end{theorem}

\begin{theorem}\label{theo:MSE}
	% Under Assumption~\ref{ass:assumption1}, 
	For a smooth test function 
	$\phi$, the MSE of an SG-MCMC with a $K$th-order integrator at time $\mathcal{T} = hT$ is bounded,
	for some $C > 0$ independent of $(T, h)$, as
	\begin{align*}
	\mathbb{E}\left(\hat{\phi} - \bar{\phi}\right)^2 \leq C \left(\frac{\frac{1}{T}\sum_t\mathbb{E}\left\|\Delta V_t\right\|^2}{T} + \frac{1}{Th} + h^{2K}\right)~.
	\end{align*}
\end{theorem}

We simplify Theorem~\ref{theo:bias1} and Theorem~\ref{theo:MSE} to Lemma~3, where the functions
$\mathcal{B}_{\text{bias}} \triangleq O\left(\frac{1}{Th} + \frac{\sum_t \left\|\mathbb{E}\Delta V_t\right\|}{T} \right)$
and $\mathcal{B}_{\text{mse}} \triangleq C \left(\frac{\frac{1}{T}\sum_t\mathbb{E}\left\|\Delta V_t\right\|^2}{T} + \frac{1}{Th}\right)$,
independent of the order of an integrator.

In addition, from Theorem~\ref{theo:bias1} and Theorem~\ref{theo:MSE}, we can get the optimal convergence rate
of a SG-MCMC algorithm with respect to the {\em Bias} and {\em MSE} by optimizing the bounds. Specifically, the 
optimal convergence rates of the {\em Bias} for the Euler integrator is $T^{-1/2}$ with optimal stepsize $\propto T^{-1/2}$, 
while this is $T^{-2/3}$ for the symmetric splitting operator with optimal stepsize $\propto T^{-1/3}$. For the MSE, the
rate for the Euler integrator is $T^{-2/3}$ with optimal stepsize $\propto T^{-1/3}$, compared to $T^{-4/5}$ with optimal
stepsize $\propto T^{-1/5}$ for the symmetric splitting integrator.

\section{Latent Dirichlet Allocation}
Following \cite{DingFBCSN:NIPS14}, this section describes the semi-collapsed posterior of the LDA model, and the {\em Expanded-Natural} representation of the prabability simplexes used in \cite{patterson2013stochastic}.

\label{ap:lda}
Let $\mathbf{W} = \{w_{j v}\}$ be the observed words, $\mathbf{Z} = \{z_{jv}\}$ be the topic indicator variables, where $j$ indexes the documents and $v$ indexes the words. Let $(\pi)_{kw}$ be the topic-word distribution, $n_{jkw}$ be the number of word $w$ in document $j$ allocated to topic $k$, $\cdot$ means marginal sum, \ie
$n_{jk\cdot} = \sum_{w} n_{dkw}$. The semi-collapsed posterior of the LDA model is
\begin{align}
p(\mathbf{W}, \mathbf{Z}, \pi | \alpha, \tau) = p(\pi | \tau) \prod_{j = 1}^J p(\wv_j, \zv_j | \alpha, \pi),
\end{align}
where $J$ is the number of documents, $\alpha$ is the parameter in the Dirichlet prior of the topic distribution for
each document, $\tau$ is the parameter in the prior of $\pi$, and
\begin{align}
p(\wv_j, \zv_j | \alpha, \pi) = \prod_{k=1}^K \frac{\Gamma\left(\alpha + n_{jk\cdot}\right)}{\Gamma\left(\alpha\right)}\prod_{w = 1}^W \pi_{kw}^{n_{jkw}}.
\end{align}

The {\em Expanded-Natural} representation of the simplexes $\pi_k$'s in \cite{patterson2013stochastic} is used, where
%which was found performed best in our algorithm though worst in \cite{Patterson:13}, let
\begin{align}
\pi_{kw} = \frac{e^{\theta_{kw}}}{\sum_{w^\prime} e^{\theta_{kw^\prime}}}~.
\end{align}
Following \cite{DingFBCSN:NIPS14}, a Gaussian prior on $\theta_{kw}$ is adopted,
\begin{align*}
p(\theta_{kw} | \tau = \{\beta, \sigma\}) = \Ncal(\theta_{kw}, \sigma^2)~.
\end{align*}
% \frac{1}{2\pi\sigma}e^{-\frac{(\theta_{kw} - \beta)^2}{2\sigma^2}}
The stochastic gradient of the log-posterior of parameter $\theta_{kw}$ with a mini-batch $\mathcal{S}_t$ becomes,
\begin{align}
& \frac{\partial \tilde{U}(\thetav)}{\partial \theta_{kw}} = \frac{\partial \log \tilde{p}(\thetav | \mathbf{W}, \tau, \alpha)}{\partial \theta_{kw}}  \\
& = \frac{\beta - \theta_{kw}}{\sigma^2}
+ \frac{J}{|\mathcal{S}_t|} \sum_{i \in \mathcal{S}_t } \mathbb{E}_{\zv_j|\wv_j, \theta, \alpha}\left(n_{jkw} - \pi_{kw}n_{jk\cdot}\right). \nonumber
\end{align}
for the $t$-th iteration, where $\mathcal{S}_t \subset \{1, 2, \cdots, J\}$, and $|\cdot|$ is the cardinality of a set.

When $\sigma = 1$, we obtain the same stochastic gradient as in \cite{patterson2013stochastic} using Riemannian manifold.
%Because our algorithm is adaptive, we can conclude our algorithm is more general than the Riemannian Langevin dynamic.
To calculate the expectation term, we use the same Gibbs sampling method as in \cite{patterson2013stochastic},
%where we Gibbs sample the documents in each batch using the following conditional distribution:
\begin{align}
p(z_{jv} = k | \wv_d, \thetav, \alpha) = \frac{\left(\alpha + n_{jk\cdot}^{\backslash v}\right)e^{\theta_{kw_{jv}}}}{\sum_{k^\prime}\left(\alpha + n_{jk^\prime\cdot}^{\backslash v}\right)e^{\theta_{k^\prime w_{jv}}}}, 
\end{align}
where $\backslash v$ denotes the count excluding the $v$-th topic assignment variable. The expectation is estimated by the samples. 

\paragraph{Running time} For the results reported in the main text, to achieve reported results of LDA on ICML dataset, mSGNHT-E, mSGNHT-S and Gibbs take 1000 iterations, the running times are 161.99, 180.55 and 516.12 second, respectively.

\begin{figure*}  \centering
	\begin{tabular}{ c  c  c  c }
		\hspace{-3mm}
		\begin{minipage}{4.3cm}
			\includegraphics[width=4.3cm]{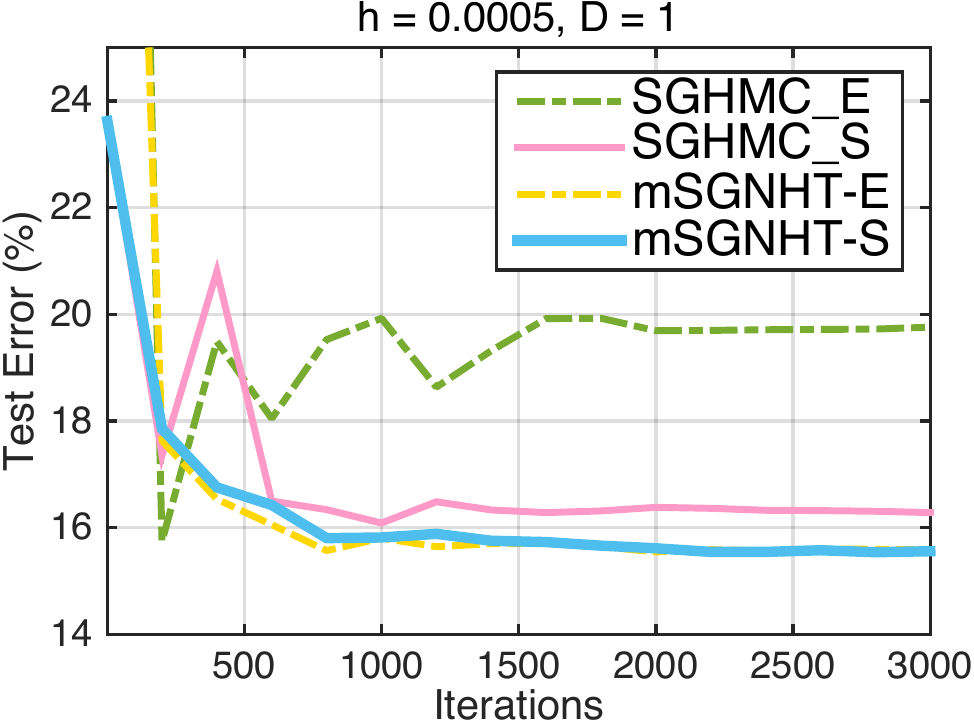} 
		\end{minipage}   &
		\hspace{-4mm}
		\begin{minipage}{4.3cm}
			\includegraphics[width=4.3cm]{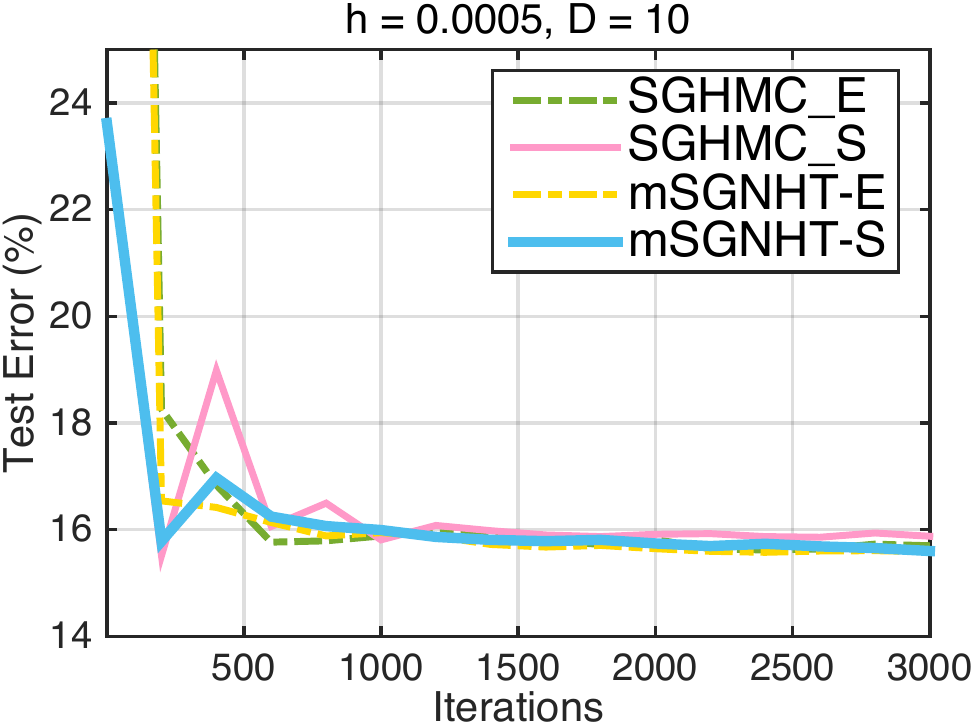} 
		\end{minipage}  &
		\hspace{-4mm}
		\begin{minipage}{4.3cm}\vspace{0mm}
			\includegraphics[width=4.3cm]{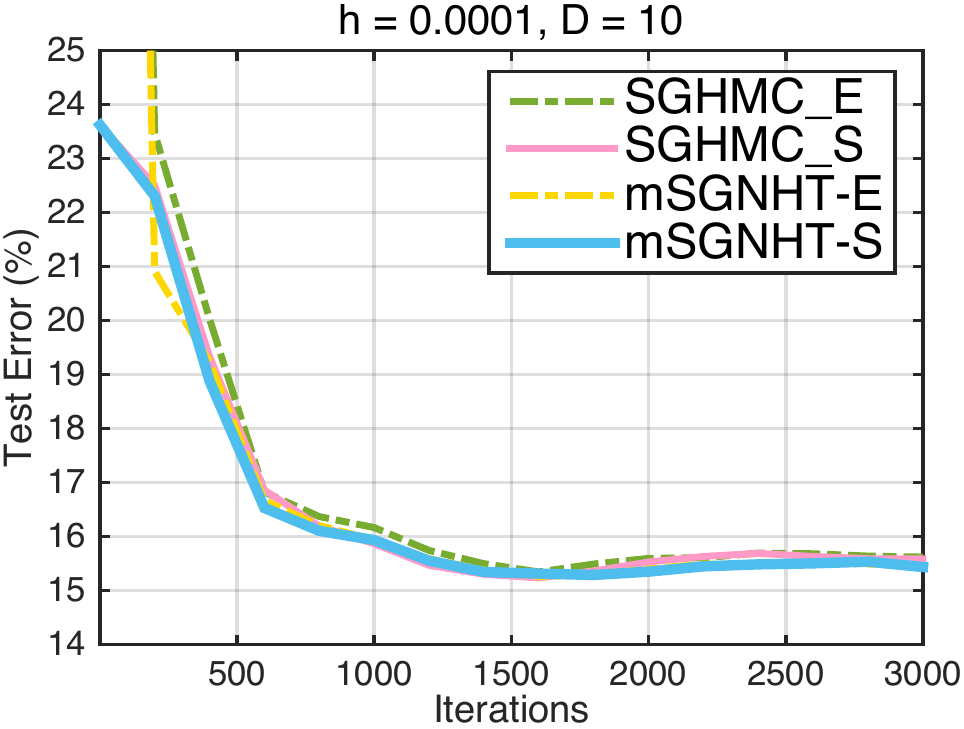} 
		\end{minipage} &
		\hspace{-4mm}
		\begin{minipage}{4.3cm}
			\includegraphics[width=4.3cm]{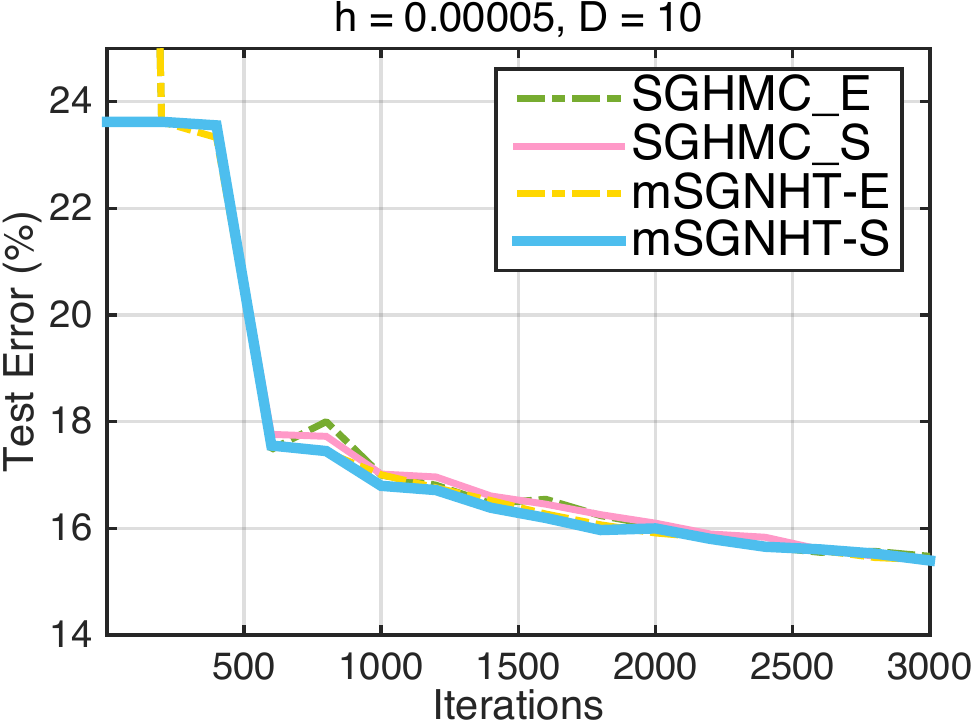} 
		\end{minipage}  \\ 
		\hspace{-3mm}
		\begin{minipage}{4.4cm}
			\includegraphics[width=4.4cm]{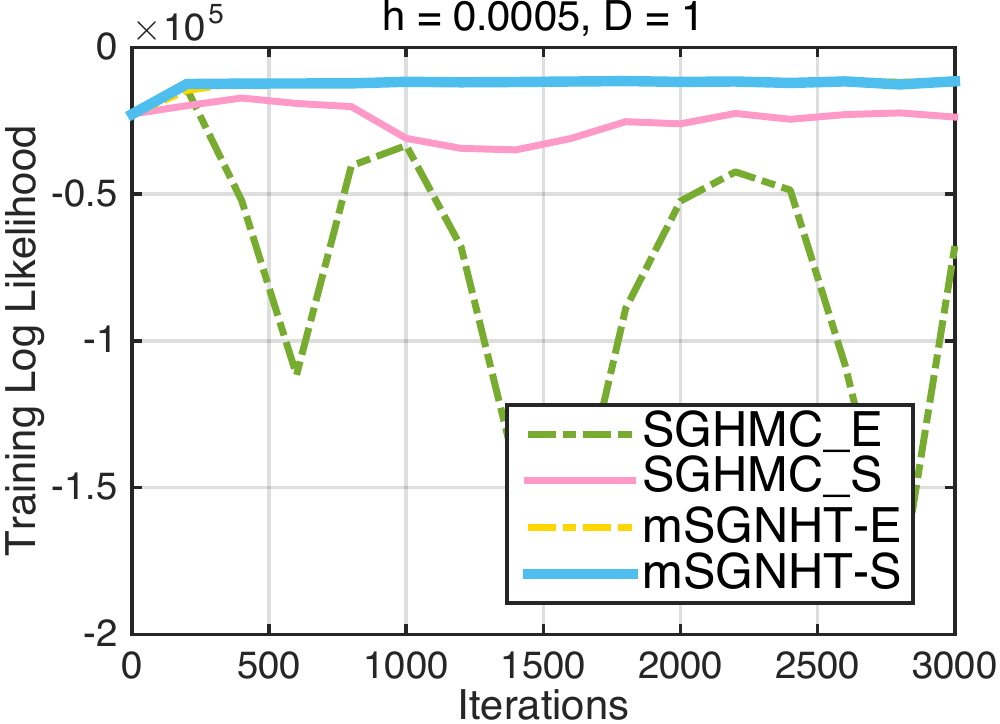} 
		\end{minipage}   &
		\hspace{-4mm}
		\begin{minipage}{4.4cm}
			\includegraphics[width=4.4cm]{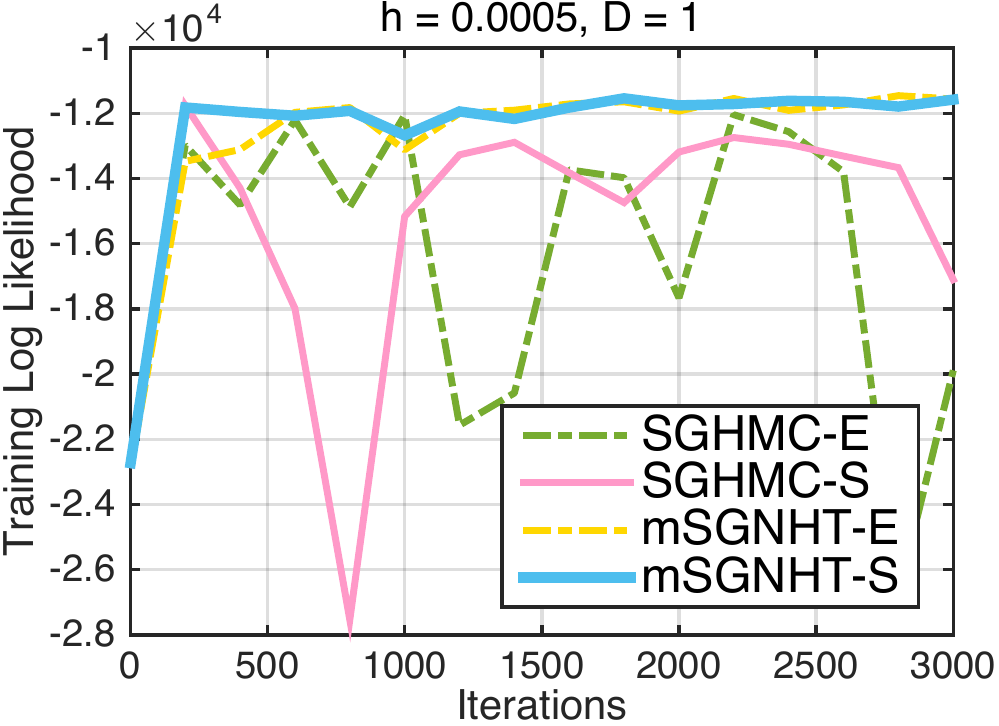} 
		\end{minipage} &
		\hspace{-4mm}
		\begin{minipage}{4.4cm}
			\includegraphics[width=4.4cm]{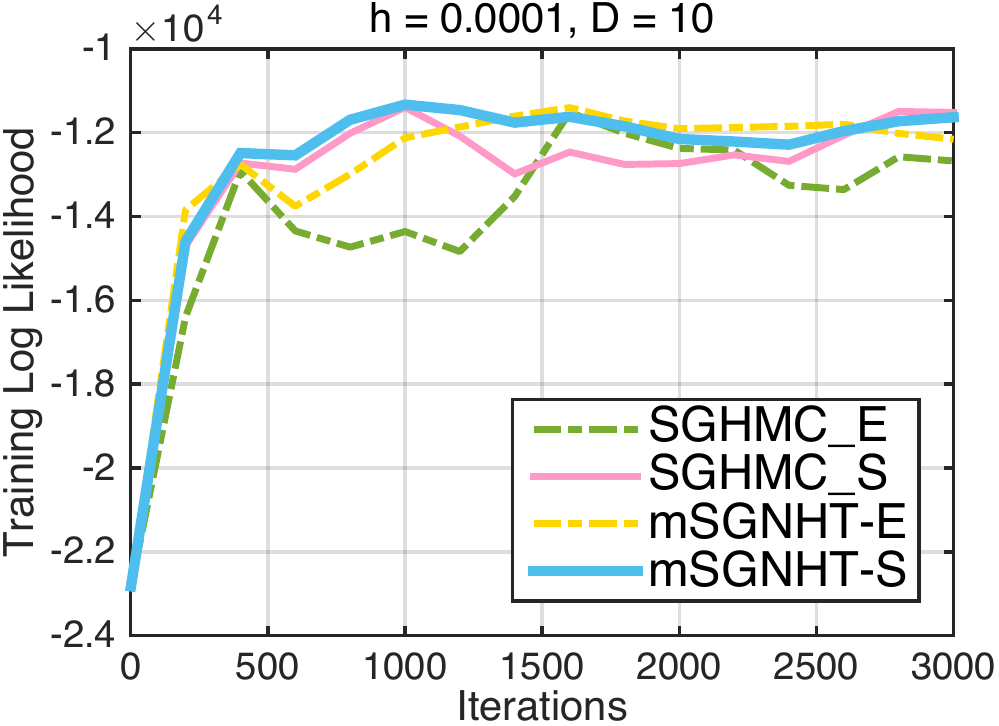} 
		\end{minipage}   &
		\hspace{-4mm}
		\begin{minipage}{4.4cm}\vspace{0mm}
			\includegraphics[width=4.4cm]{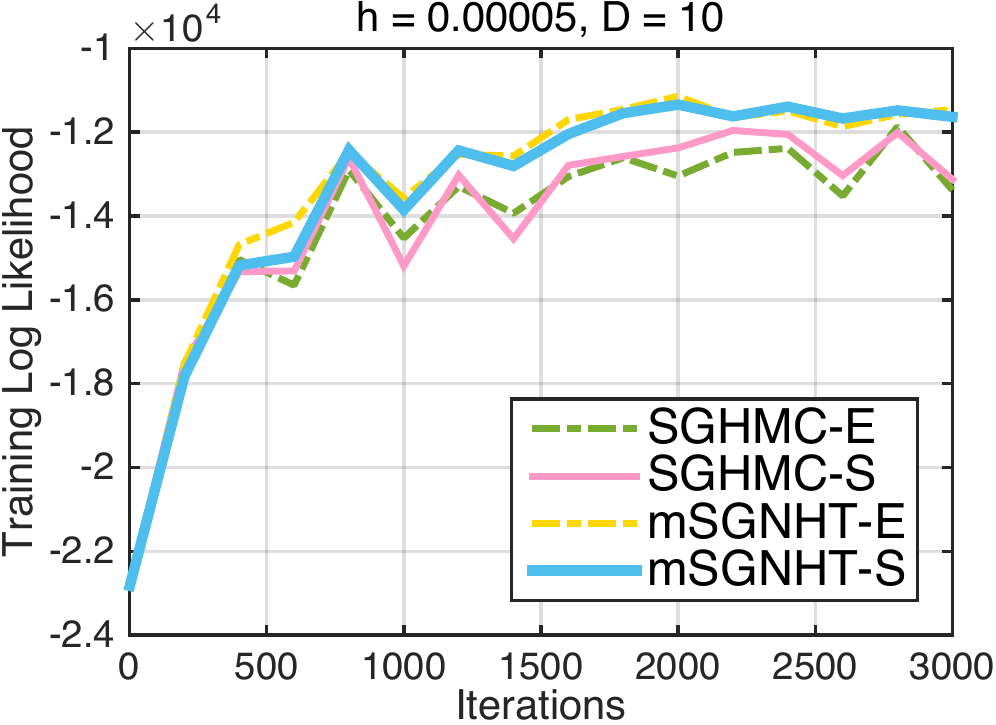} 
		\end{minipage}				
	\end{tabular}
	\caption{Learning curves for logistic regression on $\mathtt{a9a}$ dataset for varying $h$ and $D$. Column 1-2 share the same $h$; column 2-4 share the same $D$. Top row is testing error; bottom row is training log-likelihood.}		
	\label{fig:blr}	
\end{figure*}		
\vspace{-1mm}

\section{Logistic Regression}
For each data $\{\xv, y\}$, $\xv \in \R^{P}$ is the input data, $y\in \{0,1\}$ is the label. Logistic Regression gives the prabability
\begin{align}
P(y | \xv) \propto  g_{\thetav} (\xv  ) =
\frac{1}{1+\exp{ \left(  - (\Wmat^\top \xv + \cv )\right)  }}~.
\end{align}
A Gaussian prior is placed on model parameters $\thetav = \{ \Wmat, \cv\} \propto \Ncal(0, \sigma^2 \Imat)$. We set  $\sigma^2  =10$ in our experiment.

We study the effectiveness of mSGNHT-S for different $h$ and $D$. Learning curves of test errors and training log-likelihoods are shown in Fig.~\ref{fig:blr}. 
Generally, the performances of the proposed mSGNHT-S are consistently more stable than the mSGNHT-E and SGHMC, across varying $h$ and $D$. Furthermore, mSGNHT-S converges faster than mSGNHT-E, especially at the beginning of learning.
Comparing column 1-2 (fixing $h$, varying $D$),  mSGNHT-S is more robust to the choice of diffusion factor $D$.
Comparing column 2-4 (fixing $D$, varying $h$), larger $h$ potentially brings larger gradient-estimation errors and numerical errors. mSGNHT is shown to significantly outperform others when $h$ is large.

\begin{figure}[t] \tiny  \centering
	\vspace{-1mm}
	\begin{tabular}{ c  c  } % \hline
		\hspace{-0mm}
		\begin{minipage}{4.0cm}
			\includegraphics[width=4.0cm]{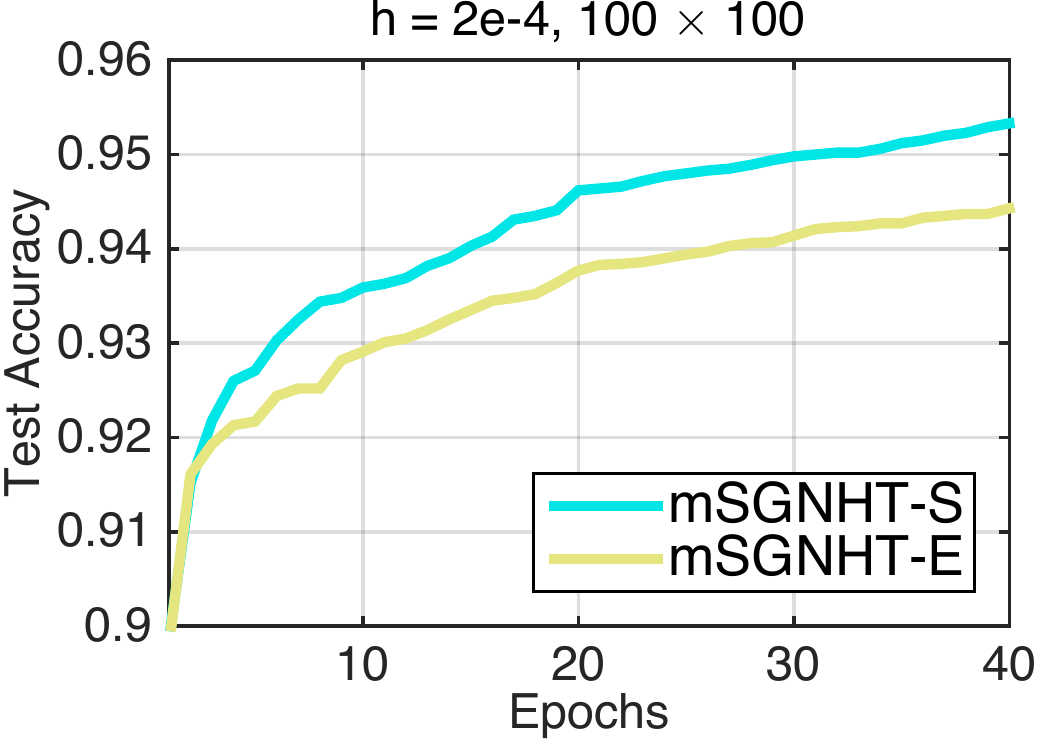} 
		\end{minipage}   &
		\hspace{-2mm}
		\begin{minipage}{4.0cm}
			\includegraphics[width=4.0cm]{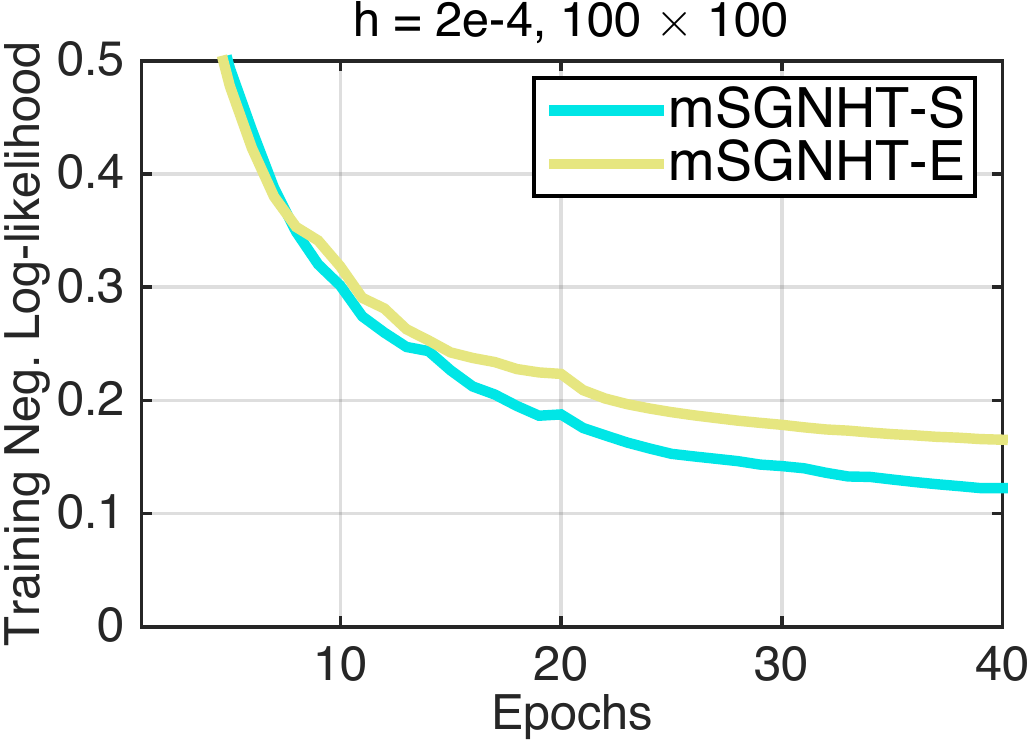} 
		\end{minipage}  \\		
		\hspace{-0mm}
		\begin{minipage}{4.0cm}
			\includegraphics[width=4.0cm]{figures/acc_results_h=0_0001_C=5_nets_100_100_10} 
		\end{minipage}  &
		\hspace{-2mm}
		\begin{minipage}{4.0cm}
			\includegraphics[width=4.0cm]{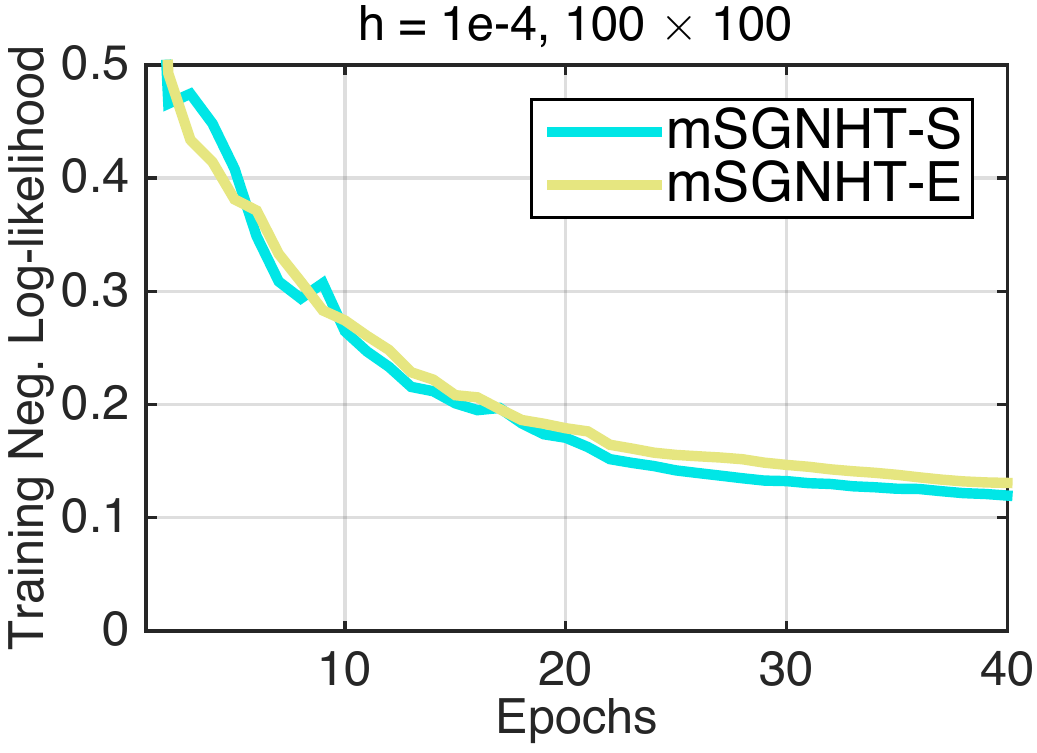} 
		\end{minipage}   \\		
		\hspace{-0mm}
		\begin{minipage}{4.0cm}\vspace{0mm}
			\includegraphics[width=4.0cm]{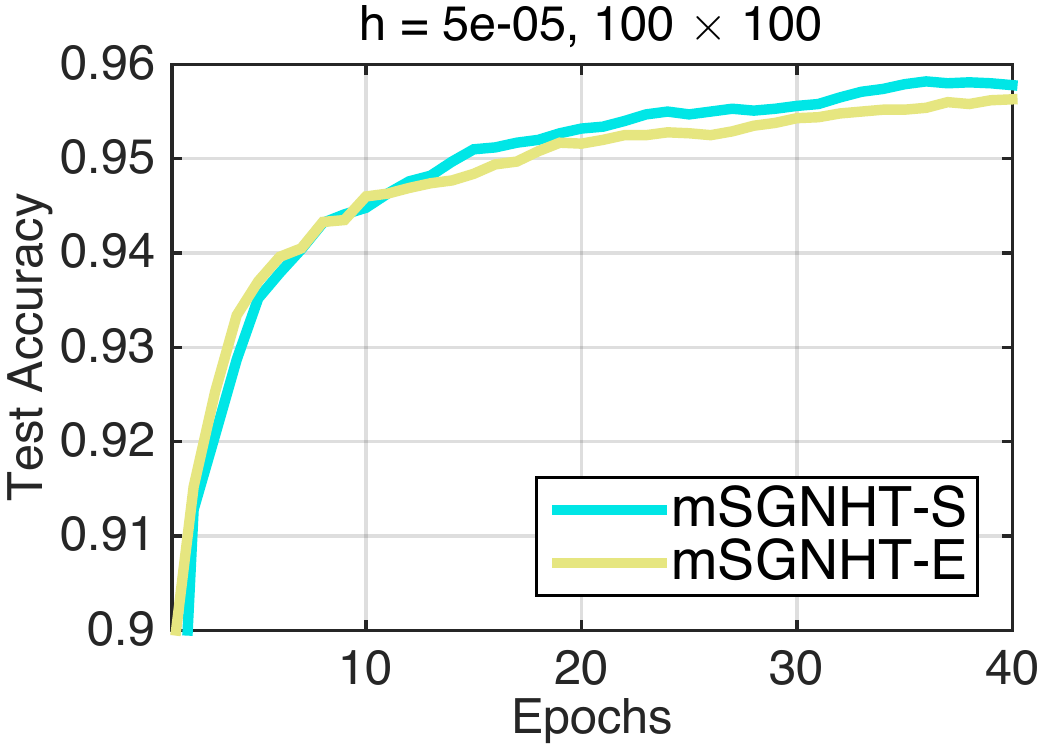} 
		\end{minipage}  &
		\hspace{-2mm}
		\begin{minipage}{4.0cm}\vspace{0mm}
			\includegraphics[width=4.0cm]{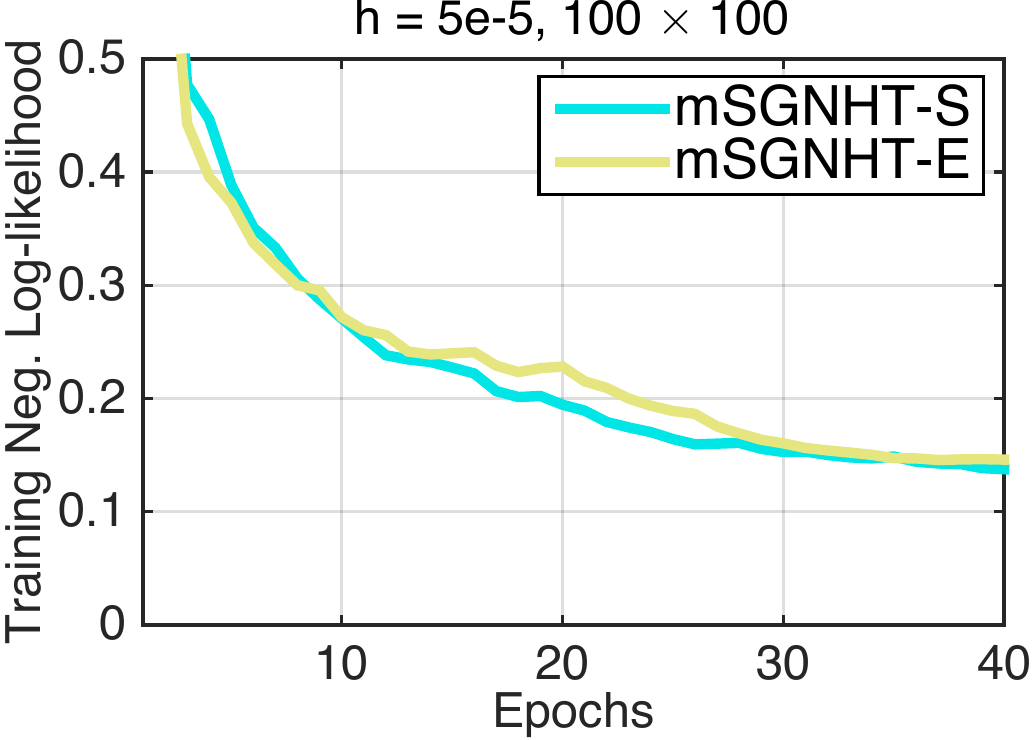} 
		\end{minipage} \\  %\hline
	\end{tabular}
	\caption{Learning curves for FNN (ReLU link) on MNIST dataset for varying stepsize $h$. Step size decreases top-down.}	
	\label{fig:fnn-stepsize}		
\end{figure}

\begin{figure}[h] \tiny  \centering
	\vspace{-2mm}
	\begin{tabular}{ c  c  } % \hline
		\hspace{-0mm}
		\begin{minipage}{3.9cm}
			\includegraphics[width=3.9cm]{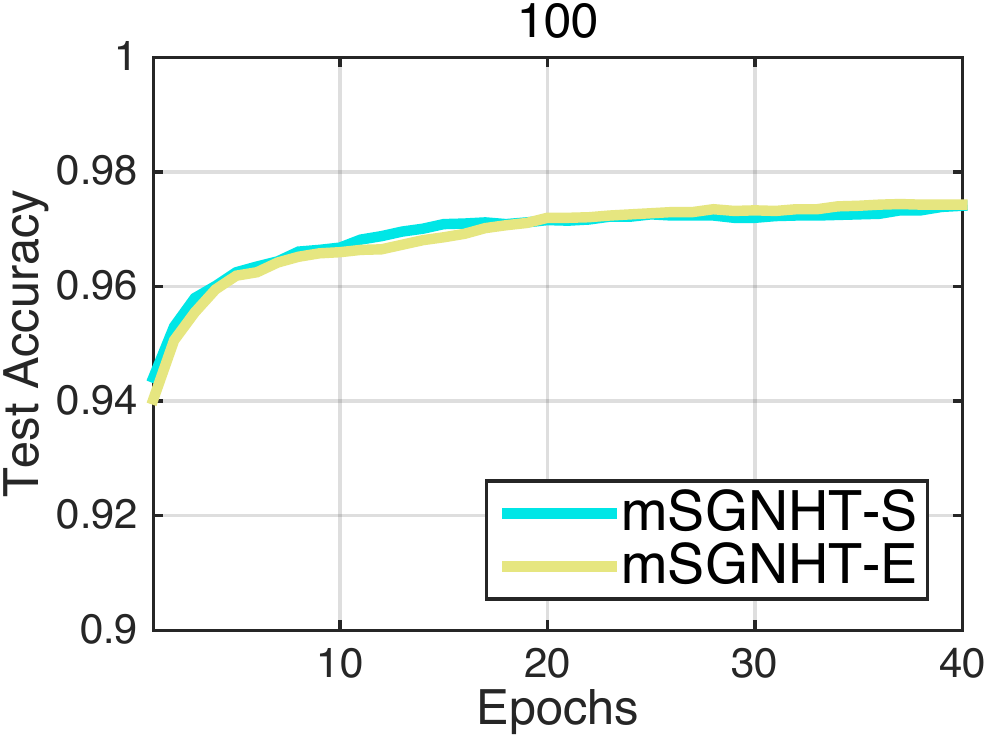} 
		\end{minipage}   &
		\hspace{-2mm}
		\begin{minipage}{3.9cm}
			\includegraphics[width=3.9cm]{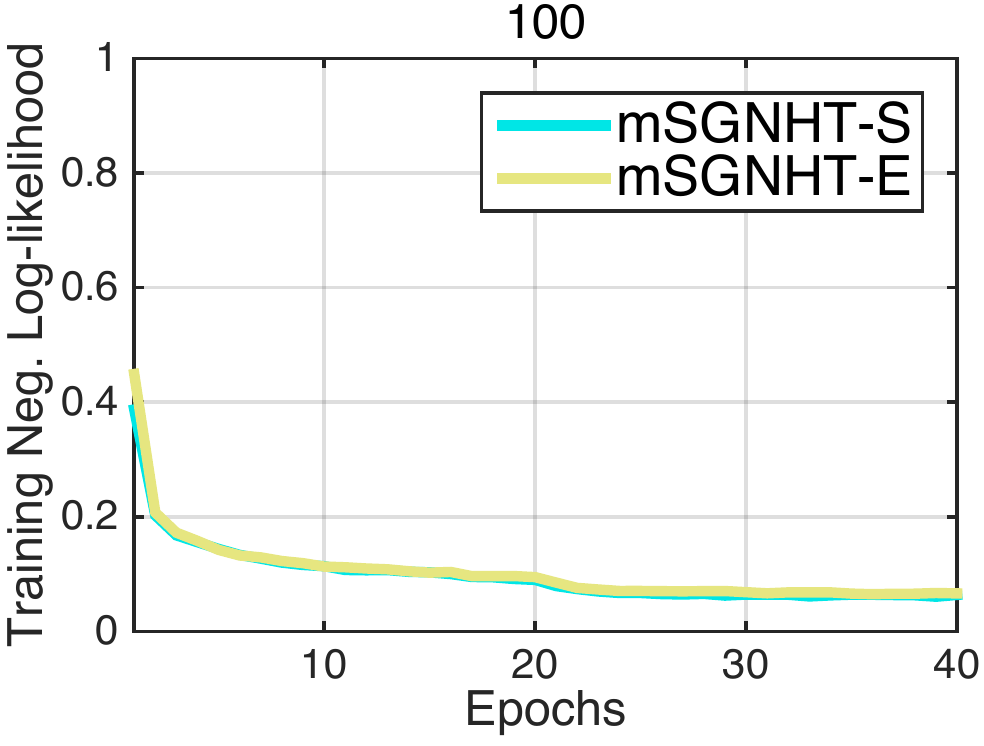} 
		\end{minipage}  \\		
		\hspace{-0mm}
		\begin{minipage}{3.9cm}
			\includegraphics[width=3.9cm]{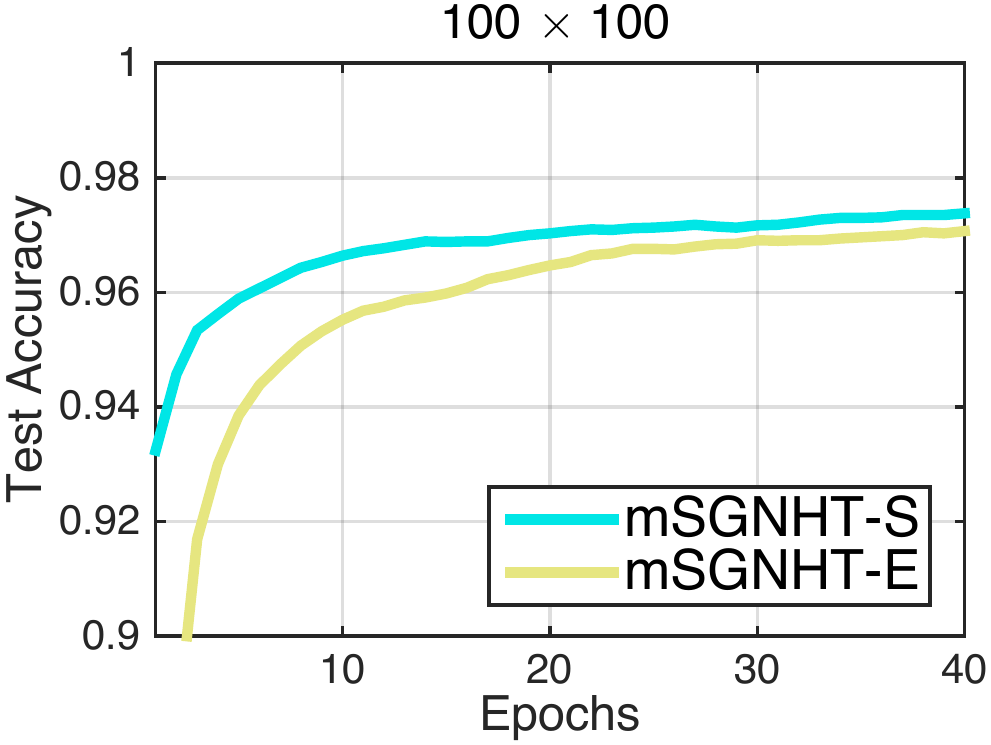} 
		\end{minipage}  &
		\hspace{-2mm}
		\begin{minipage}{3.9cm}
			\includegraphics[width=3.9cm]{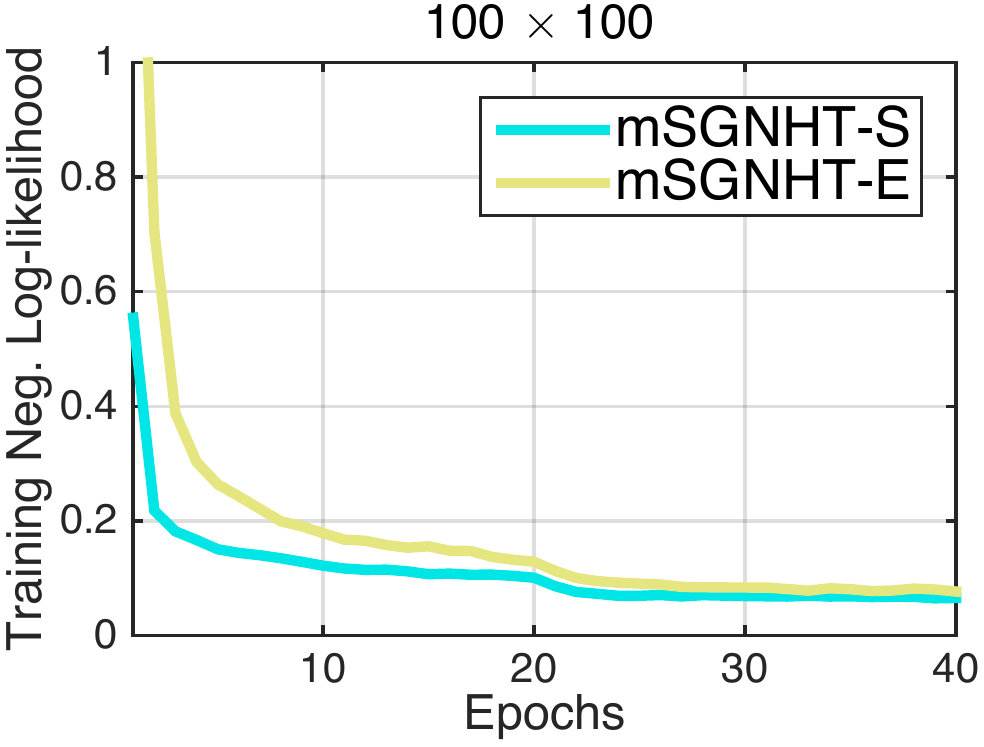} 
		\end{minipage}   \\		
		\hspace{-0mm}
		\begin{minipage}{3.9cm}\vspace{0mm}
			\includegraphics[width=3.9cm]{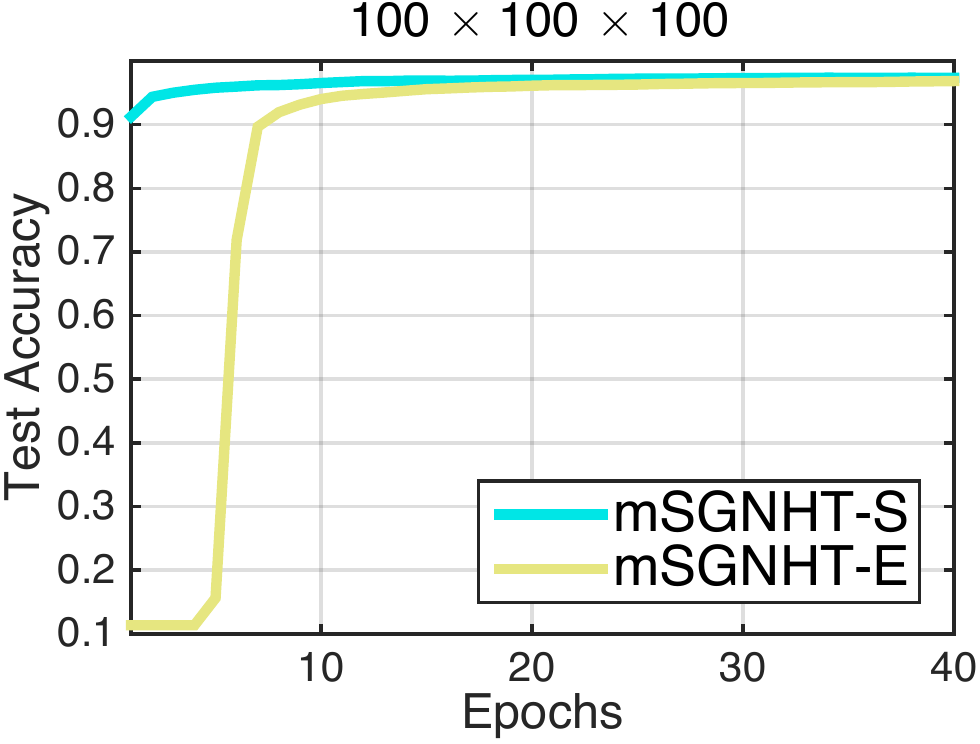} 
		\end{minipage}  &
		\hspace{-2mm}
		\begin{minipage}{3.9cm}\vspace{0mm}
			\includegraphics[width=3.9cm]{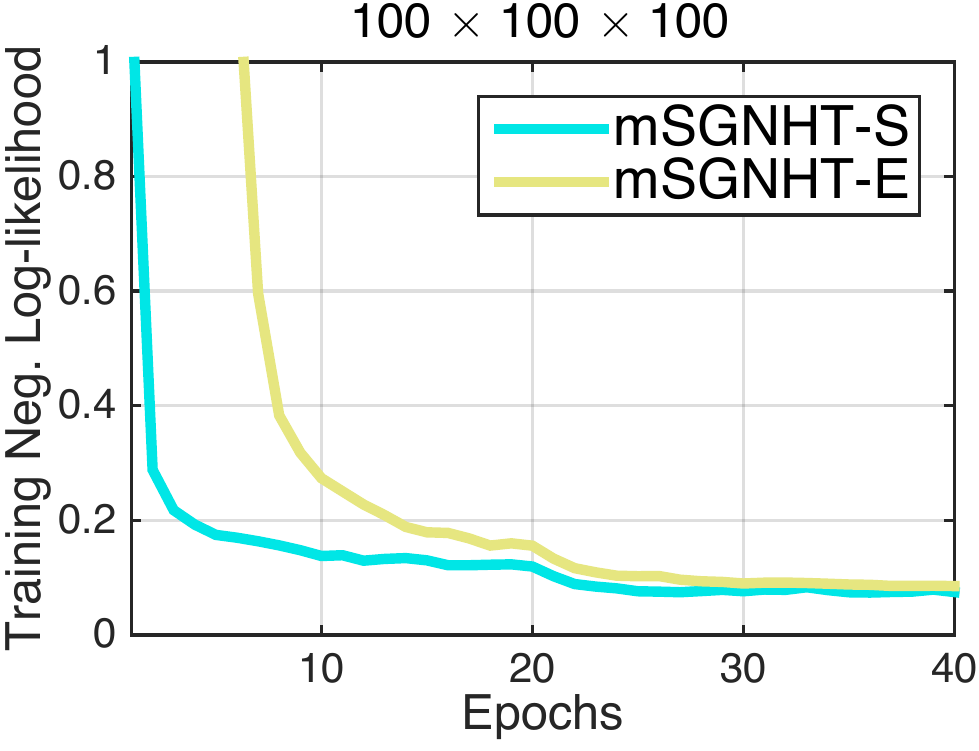} 
		\end{minipage} \\  %\hline
	\end{tabular}
	\vspace{-2mm}
	\caption{Learning curves for FNN (Sigmoid link) on MNIST dataset for varying network depth. Depth increases top-down.}	
	\vspace{-2mm}
	\label{fig:fnn-sigmoid}		
\end{figure}		

\section{Deep Neural Networks}

\subsection{Sensitivity to Step-size}\vspace{-1mm}
For feedforward neural nets (FNN) with a 2-layer model of 100-100 ReLU, we study the performance of mSGNHT-S for different $h$. $D=5$. We test a wide range of $h$, and show  in Fig.~\ref{fig:fnn-stepsize} the learning curves of the test accuracy and training negative log-likelihood for $h=2\!\times\!10^{-4}, 1\!\times\!10^{-4}, 5\!\times\!10^{-5} $. mSGNHT-S is less sensitive to step size; it maintains fast convergence when $h$ is large, while mSGNHT-S significantly slows down.

\subsection{Sigmoid Activation Function}
We compare different methods in the case of sigmoid link. Similar to the main paper, we test the FNNs with varying depths, \eg $\{1, 2, 3\}$, respectively. We set $D=10$ and $h=10^{-3}$. Fig.~\ref{fig:fnn-sigmoid} displays learning curves on testing accuracy and training negative log-likelihood. The gaps between mSGNHT-S and mSGNHT-E becomes larger in deeper models. When a 3-layer network is emplyed, mSGNHT-E fails at the first 5 epochs, whilst mSGNHT-S converges pretty well. Moreover, mSGNHT-S converges more accurately, yielding an accuracy $97.36\%$, while mSGNHT-E gives $96.86\%$. This manifests the importance of numerical accuracy in SG-MCMC for practical applications, and mSGNHT-S is desirable in this regard.

\subsection{Comparison with Other Methods}
Based on network size 400-400 with ReLU, we also compared with a recent state-of-the-art inference method for neural nets, Bayes by Backprop (BBB)~\cite{blundell2015weight}. $h=2\times10^{-4}$ and $D=60$. The comparison is in Table~\ref{tab:fnn}.
BBB and SGD are results taken from~\cite{blundell2015weight}.

\begin{table}[h]\centering
	\caption{Classification accuracy on MNIST.}
	\label{tab:fnn}
\begin{tabular}{c  c}
	\hline
	% \abovespace \belowspace
	{\bf Method} & {\bf Accuracy (\%) $\uparrow$} \\
	\hline 
	mSGNHT-S     &  {\bf 98.25}\\
	mSGNHT-E				     &  98.20\\	
%	SGLD         &  98.25\\		  
	\hline			
	BBB              &  98.18\\		 	
	SGD			    &  98.17\\		 	
	\hline 
\end{tabular}
\end{table}

\subsection{Convolutional Neural Networks}
\begin{figure}[t!] \centering
	\begin{tabular}{c c}
		\hspace{-4mm}
		\includegraphics[width=4.2cm]{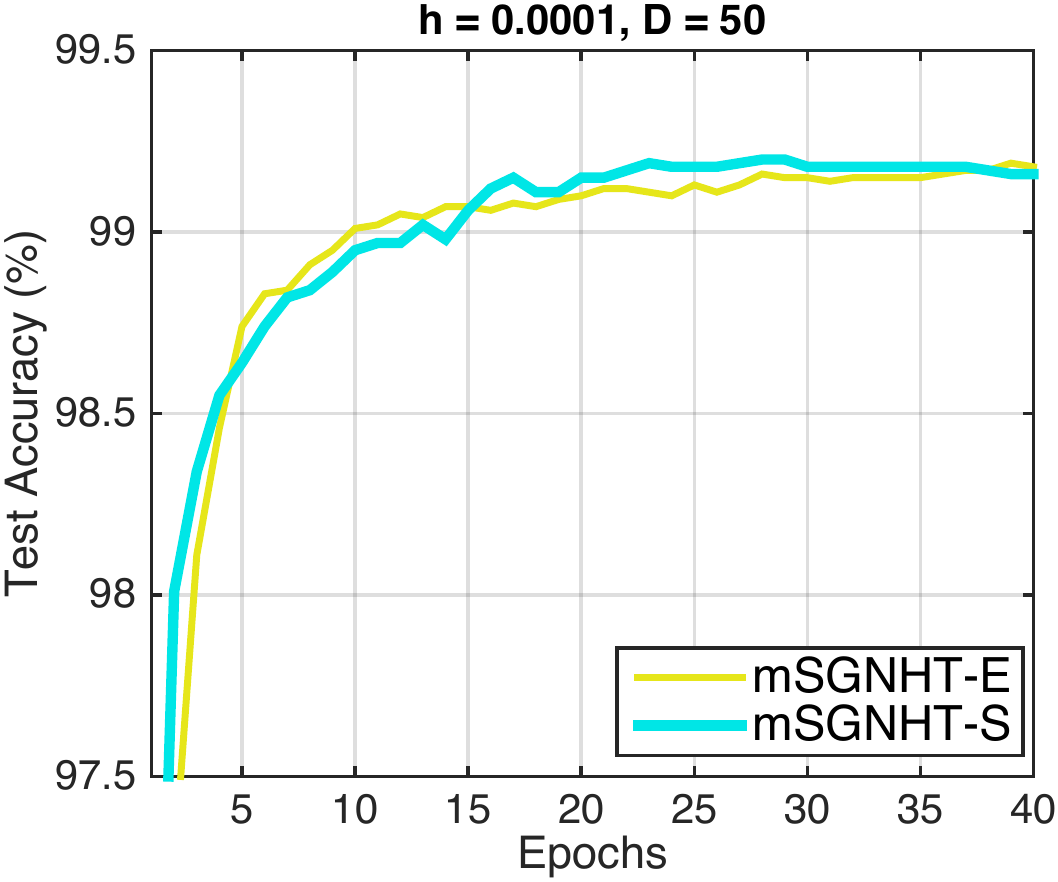} &
		\hspace{-4mm}
		\includegraphics[width=4.2cm]{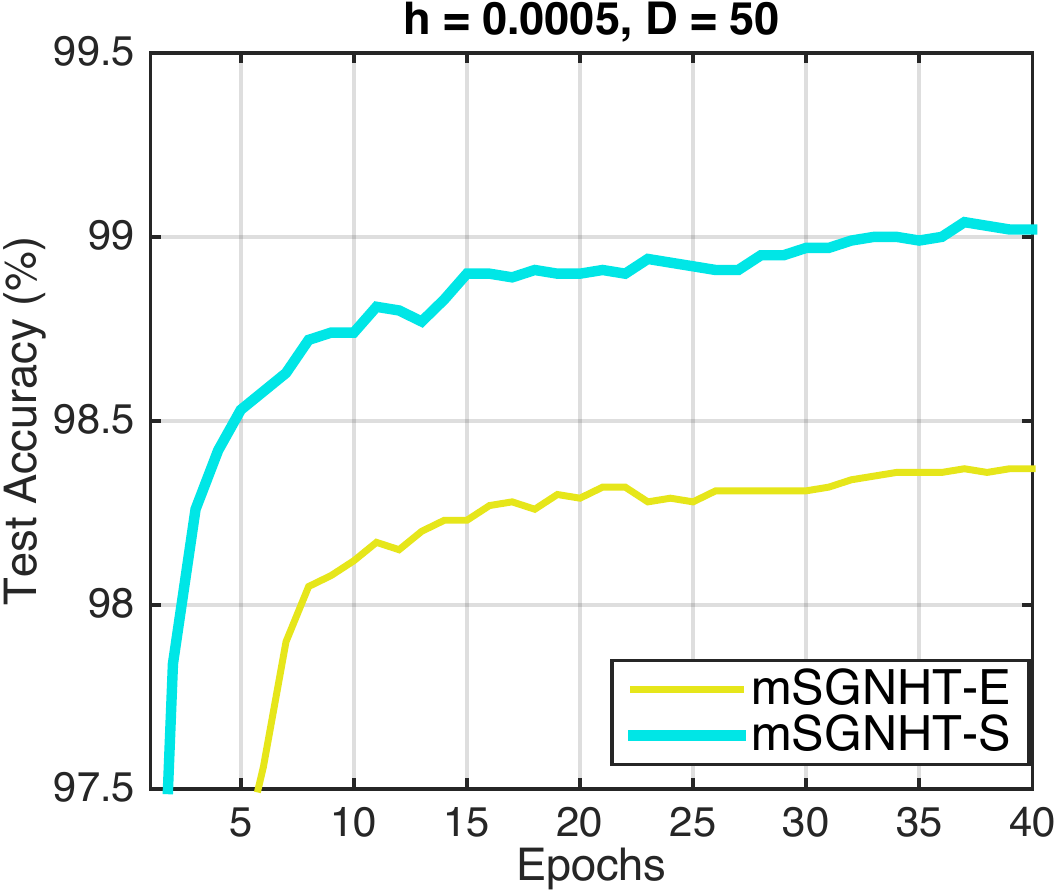} \\
%		(a) 400-400 & (b) 800-800
	\end{tabular} \vspace{-1mm}
	\caption{Learning curves of CNN for different step sizes.}
	\label{fig:cnn}
	\vspace{0pt}
\end{figure}

We also performed the comparison with a standard network convolutional neural networks, LeNet~\cite{jarrett2009best}, on MNIST dataset. It is 2-layer convolution networks followed by a 2-layer fully-connected FNN, each containing 200 hidden nodes that uses ReLU. Both convolutional layers use $5 \times 5$ filter size with 32 and 64 channels, respectively, $2 \times 2$ max pooling are used after each convolutional layer. 40 epochs are used, and $L$ is set to 20. In Fig.~\ref{fig:cnn}, we tested the stepsizes $h=10^{-4}$ and $h=5\times10^{-4}$ for mSGNHT-S and mSGNHT-S, and $D=50$.

Again, under the same network architecture, CNN trained with mSGNHT-S converges fater than mSGNHT-E. In particular, when a large stepsize used, mSGNHT-S has a significant improvement over mSGNHT-E.

\section{Deep Poisson Factor Analysis}

\subsection{Model Specification}\vspace{-1mm}
We first provide model details of Deep Poisson Factor Analysis (DPFA)~\cite{gan2015scalable}.
Given a discrete matrix $\Wmat \in \Z_+^{V \times J}$ containing counts from
$J$ documents and $V$ words, Poisson factor analysis (PFA)~\cite{zhou2015negative} assumes the entries of $\Wmat$ are summations of $K < \infty$ latent counts, each produced by a latent factor (in the case of topic modeling, a hidden topic). For $\Wmat$, the generative process of DPFA with $L$-layer Sigmoid Belief Networks (SBN), is as follows
\begin{align}
p(h_{k_{L}}^{(L)}) &  = g(b_{k_{L}}^{(L)}) \label{eq:sbn1}\\
p(h_{k_{\ell}n}^{(\ell)} =1  | \hv_{n}^{(\ell+1)}  )& = g \big( ( \wv_{k_{\ell}}^{(\ell)} )^\top \hv_{n}^{(\ell+1)}  + c_{k_{\ell}}^{(\ell)} \big) \label{eq:sbn2}\\
\Wmat &\sim \Pois(\Phimat (\Psimat   \odot \hv^{(1)} ))
\end{align} 
where $g(\cdot)$ is the Sigmoid link. Equation~(\ref{eq:sbn1}) and (\ref{eq:sbn2}) define Deep Sigmoid Belief Networks (DSBN). $\Phimat$ is the factor loading matrix. Each column of $\Phimat$,
$\phiv_{k} \in \Delta_V$ , encodes the relative importance of each word in
topic $k$, with $\Delta_V$ representing the $V$-dimensional simplex.
$\Thetamat \in \R_+^{K \times J}$  is the factor score matrix. Each column $\psiv_{j}$,
contains relative topic intensities specific to document $j$. $\hv^{(\ell)} \in \{0,1\}^{K \times 1}$ is a latent binary feature vector, which defines whether certain topics are associated with documents.  The factor scores for document $j$ at bottom layer are the element-wise multiplication $\psiv_{j}  \odot \hv^{(1)}$. DPFA is constructed by placing Dirichlet priors on $\Phimat_k$ and gamma priors on $\psiv_{j}$. This is summarized as,
\begin{align}
x_{vj} = \sum_{k=1}^{K} x_{vjk}, \qquad x_{vjk} 
\sim \Pois(\phi_{vk} \psi_{kj}   h_{k}  )
\end{align} 
with priors specified as 
$\phiv_{k} \sim \Dir(a_\phi, \dots, a_\phi )$, 
$\psi_{kn} \sim \Gal(r_{k}, p_k/(1-p_k) )$,
$r_{k} \sim \Gal(\gamma_{0}, 1/c_0  )$,
and 
$\gamma_{0} \sim \Gal(e_0, 1/f_0 )$.

\subsection{Model Inference}\vspace{-1mm}
Following~\cite{gan2015scalable}, we use stochastic gradient Riemannian Langevin dynamics (SGRLD) \cite{patterson2013stochastic} to sample the topic-word distributions $\{\phiv_k\}$. mSGNHT is used to sample the parameters in DSBN, \ie $\thetav = \left( \Wmat^{(\ell)} , \cv^{(\ell)}, \bv^{(L)} \right)$, where $\ell = 1, \cdots, L-1$.
Specifically, the stochastic gradients of $\Wmat^{(\ell)}$ and $\cv^{(\ell)}$ evaluated on a mini-batch of data (denote $\mathcal{S}$ as the index set of a mini-batch) are calculated,
\begin{align}
\frac{\partial \tilde{U}}{\partial \wv_{k_{\ell}}^{(\ell)}} = & \ \frac{J}{|\mathcal{S}|}\sum_{i \in \mathcal{D}} \mathbb{E}_{\hv_i^{(\ell)}, \hv_i^{(\ell+1)}}\left[\left(\tilde{\sigma}_{k_{\ell}i}^{(\ell)} - h_{k_{\ell}i}^{(\ell)} \right) \hv_{i}^{(\ell+1)}\right] \,, \\
\frac{\partial \tilde{U}}{\partial c_{k_{\ell}}^{(\ell)}} = & \ \frac{J}{|\mathcal{S}|}\sum_{i \in \mathcal{D}} \mathbb{E}_{\hv_i^{(\ell)}, \hv_i^{(\ell+1)}}\left[\tilde{\sigma}_{k_{\ell}i}^{(\ell)} - h_{k_{\ell}i}^{(\ell)}\right] \,,
\end{align}
where $\tilde{\sigma}_{k_{\ell}i}^{(\ell)} = \sigma((\wv_{k_{\ell}}^{(\ell)})^\top \hv_{i}^{(\ell+1)} + c_{k_{\ell}}^{(\ell)})$, and the expectation is taken over posteriors. Monte Carlo integration is used to approximate this quantity.

% \newpage
% \bibliographystyle{aaai}	
% \bibliography{subtex/references.bib}

% \end{document}